%% file: main.tex
\documentclass[onecolumn,journal,11pt]{IEEEtran}

\usepackage{epsfig}
\usepackage{graphicx}
\usepackage{caption}
\usepackage{amsmath,amsthm}
\usepackage{amssymb}
\usepackage{mathrsfs} 
\usepackage{enumerate}
\usepackage{multirow}
\usepackage{tabularx}
\usepackage{float}
\usepackage[numbers,sort]{natbib}
\usepackage{color}
\usepackage{times}
\usepackage{algorithm}%
\usepackage{algorithmic}
\usepackage[breaklinks=true,bookmarks=false]{hyperref}

\input{iPythonNotebookConfig.tex}

\title{Sliced-Wasserstein Autoencoder: An Embarrassingly Simple Generative Model}

\author{\IEEEauthorblockN{Soheil Kolouri\IEEEauthorrefmark{1}, Phillip E. Pope\IEEEauthorrefmark{1}, Charles E. Martin\IEEEauthorrefmark{1}, Gustavo K. Rohde\IEEEauthorrefmark{2}}
 \\ \IEEEauthorblockA{\IEEEauthorrefmark{1}{  HRL Laboratories, LLC,  Malibu, CA 91302}\\
 \IEEEauthorrefmark{2}{University of Virginia, Charlottesville, VA 22908}} 
\\ {\tt\small skolouri@hrl.com, cemartin@hrl.com, gustavo@virginia.edu}
}

\begin{document}

\maketitle

\begin{abstract}
In this paper we study generative modeling via autoencoders while using the elegant geometric properties of the optimal transport (OT) problem and the Wasserstein distances. We introduce Sliced-Wasserstein Autoencoders (SWAE), which are generative models that enable one to shape the distribution of the latent space into any samplable probability distribution without the need for training an adversarial network or defining a closed-form for the distribution. In short, we regularize the autoencoder loss with the sliced-Wasserstein distance between the distribution of the encoded training samples and a predefined samplable distribution. We show that the proposed formulation has an efficient numerical solution that provides similar capabilities to Wasserstein Autoencoders (WAE) and Variational Autoencoders (VAE), while benefiting from an embarrassingly simple implementation. 
\end{abstract}

\section{Introduction}

Scalable generative models that capture the rich and often nonlinear distribution of high-dimensional data, (i.e., image, video, and audio), play a central role in various applications of machine learning, including transfer learning \cite{isola2017image,murez2017image}, super-resolution \cite{ledig2016photo,kolouri2015transport}, image inpainting and completion \cite{yeh2017semantic}, and image retrieval \cite{creswell2016adversarial}, among many others. 
The recent generative models, including Generative Adversarial Networks (GANs) \cite{goodfellow2014generative,radford2015unsupervised,arjovsky2017wasserstein,berthelot2017began} and  Variational Autoencoders (VAE) \cite{kingma2013auto,mescheder2017adversarial,bousquet2017optimal} enable an unsupervised and end-to-end modeling of the high-dimensional distribution of the training data.  

Learning such generative models boils down to minimizing a dissimilarity measure between the data distribution and the output distribution of the generative model. To this end, and following the work of \citet{arjovsky2017wasserstein} and \citet{bousquet2017optimal} we approach the problem of generative modeling from the optimal transport point of view. The optimal transport problem \cite{villani2008optimal,kolouri2017optimal} provides a way to measure the distances between probability distributions by transporting (i.e., morphing) one distribution into another. Moreover, and as opposed to the common information theoretic dissimilarity measures (e.g., $f$-divergences), the p-Wasserstein dissimilarity measures that arise from the optimal transport problem: 1) are true distances, and 2) metrize a weak convergence of probability measures (at least on compact spaces). Wasserstein distances have recently attracted a lot of interest in the learning community \cite{frogner2015learning,gulrajani2017improved,bousquet2017optimal,arjovsky2017wasserstein,kolouri2017optimal} due to their exquisite geometric characteristics \cite{santambrogio2015optimal}. 
See the supplementary material for an intuitive example showing the benefit of the Wasserstein distance over commonly used $f$-divergences. 

In this paper, we introduce a new type of autoencoders for generative modeling (Algorithm 1), which we call Sliced-Wasserstein Autoencoders (SWAE), that minimize the sliced-Wasserstein distance between the distribution of the encoded samples and a predefined samplable distribution.  Our work is most closely related to the recent work by \citet{bousquet2017optimal} and the follow-up work by \citet{tolstikhin2017wasserstein}. However, our approach avoids the need to perform costly adversarial training in the encoding space and is not restricted to closed-form distributions, while still benefiting from a Wasserstein-like distance measure in the encoding space that permits a simple numerical solution to the problem.

In what follows we first provide an extensive review of the preliminary concepts that are needed for our formulation. In Section 3 we formulate our proposed method. The proposed numerical scheme to solve the problem is presented in Section 4. Our experiments are summarized in Section 5. Finally, our work is Concluded in Section 6.

\section{Notation and Preliminaries}

Let $X$ denote the compact domain of a manifold in Euclidean space and let $x_n\in X$ denote an individual input data point. Furthermore, let $\rho_X$ be a Borel probability measure defined on $X$. We define the probability density function $p_X(x)$ for input data $x$ to be:
\begin{equation*}
d\rho_X(x)=p_X(x)dx
\end{equation*}

Let $\phi:X\rightarrow Z$ denote a deterministic parametric mapping from the input space to a latent space $Z$ (e.g., a neural network encoder). Utilizing a technique often used in the theoretical physics community (See \cite{gillespie1983theorem}), known as Random Variable Transformation (RVT), the probability density function of the encoded samples $z$ can be expressed in terms of $\phi$ and $p_X$ by:
\begin{equation}
p_Z(z)=\int_X p_X(x)\delta(z-\phi(x))dx,
\label{eq:rvt}
\end{equation}
where $\delta$ denotes the Dirac distribution function. The main objective of Variational Auto-Encoders (VAEs) is to encode the input data points $x\in X$ into latent codes $z\in Z$ such that: 1) $x$ can be recovered/approximated from $z$, and 2)  the probability density function of the encoded samples, $p_Z$, follows a prior distribution $q_Z$. Similar to classic auto-encoders, a decoder $\psi:Z\rightarrow X$ is required to map the latent codes back to the original space such that 
\begin{equation}
p_{Y}(y)=\int_Xp_X(x)\delta(y-\psi(\phi(x)))dx,
\label{eq:py}
\end{equation}
where $y$ denotes the decoded samples. It is straightforward to see that when $\psi=\phi^{-1}$ (i.e. $\psi(\phi(\cdot))=id(\cdot)$), the distribution of the decoder $p_Y$ and the input distribution $p_X$ are identical. Hence, the objective of a variational auto-encoder simplifies to learning $\phi$ and $\psi$ such that they minimize a dissimilarity measure between $p_Y$ and $p_X$, and between $p_Z$ and $q_Z$. Defining and implementing the dissimilarity measure is a key design decision, and is one of the main contributions of this work,  and thus we dedicate the next section to describing existing methods for measuring these dissimilarities. 

\subsection{Dissimilarity between $p_X$ and $p_Y$}

We first emphasize that the VAE work in the literature often assumes stochastic encoders and decoders \cite{kingma2013auto}, while we consider the case of only deterministic mappings.
Different dissimilarity measures have been used between $p_X$ and $p_Y$ in various work in the literature.  Most notably, \citet{nowozin2016f} showed that for the general family of $f$-divergences, $D_f(p_X,p_Y)$, (including the KL-divergence, Jensen-Shannon, etc.), using the Fenchel conjugate of the convex function $f$ and minimizing $D_f(p_X,p_Y)$ leads to a min-max problem that is equivalent to the {\it adversarial training} widely used in the generative modeling literature \cite{goodfellow2014generative,makhzani2015adversarial,mescheder2017adversarial}. 

Others have utilized the rich mathematical foundation of the OT problem and Wasserstein distances \cite{arjovsky2017wasserstein,gulrajani2017improved,bousquet2017optimal, tolstikhin2017wasserstein}.   In Wasserstein-GAN, \cite{arjovsky2017wasserstein} utilized the Kantorovich-Rubinstein duality for the 1-Wasserstein distance, $W_1(p_X,p_Y)$, and reformulated the problem as a min-max optimization that is solved through an adversarial training scheme. In a different approach, \cite{bousquet2017optimal} utilized the autoencoding nature of the problem and showed that $W_c(p_X,p_Y)$ could be simplified as:
\begin{equation}
W_c(p_X,p_Y)= \int_X p_X(x)c(x,\psi(\phi(x)))dx
\label{eq:wc}
\end{equation}
Note that Eq. \eqref{eq:wc} is equivalent to Theorem 1 in \cite{bousquet2017optimal} for deterministic encoder-decoder pair, and also note that $\phi$ and $\psi$ are parametric differentiable models (e.g. neural networks). 
Furthermore, Eq. \eqref{eq:wc} supports a simple implementation where for i.i.d samples of the input distribution $\{x_n\}_{n=1}^N$ the minimization can be written as:
\begin{equation}
W_c(p_X,p_Y)=\frac{1}{N}\sum_{n=1}^N c(x_n,\psi(\phi(x_n)))
\label{eq:wcsimple}
\end{equation}

We emphasize that Eq. \eqref{eq:wc} (and consequently Eq. \eqref{eq:wcsimple}) takes advantage of the fact that the pairs $x_n$ and $y_n=\psi(\phi(x_n))$ are available, hence calculating the transport distance coincides with summing the transportation costs between all pairs ($x_n$, $y_n$). For example, the total transport distance may be defined as the sum of Euclidean distances between all pairs of points. In this paper, we also use $W_c(p_X,p_Y)$ following Eq. \eqref{eq:wcsimple} to measure the discrepancy between $p_X$ and $p_Y$. Next, we review the methods used for measuring the discrepancy between $p_Z$ and $q_Z$. 

\subsection{Dissimilarity between $p_Z$ and $q_Z$}

If $q_Z$ is a known distribution with an explicit formulation (e.g. Normal distribution) the most straightforward approach for measuring the (dis)similarity between $p_Z$ and $q_Z$ is the log-likelihood of $z=\phi(x)$ with respect to $q_Z$, formally:
\begin{equation}
sup_{\phi} \int_X p_X(x)log(q_Z(\phi(x)))dx
\label{eq:maxll}
\end{equation}
maximizing the log-likelihood is equivalent to minimizing the KL-divergence between $p_Z$ and $q_Z$, $D_{KL}(p_Z,q_Z)$ (see supplementary material for more details and derivation of Equation \eqref{eq:maxll}). This approach has two major limitations: 1) The KL-Divergence and in general $f$-divergences do not provide meaningful dissimilarity measures for distributions supported on non-overlapping low-dimensional manifolds \cite{arjovsky2017wasserstein,kolouri2017sliced} (see supplementary material), which is common in hidden layers of neural networks, and therefore they do not provide informative gradients for training $\phi$, and 2) we are limited to distributions $q_Z$ that have known explicit formulations, which is very restrictive because it eliminates the ability to use the much broader class of distributions were we know how to sample from them, but do not know their explicit form. 

Various alternatives exist in the literature to address the above-mentioned limitations. These methods often sample $\tilde{\mathcal{Z}}=\{\tilde{z}_j\}_{j=1}^N$ from $q_Z$ and $\mathcal{Z}=\{z_n=\phi(x_n)\}_{n=1}^N$ from $p_X$ and measure the discrepancy between these sets (i.e. point clouds). Note that there are no one-to-one correspondences between $\tilde{z}_j$s and $z_n$s. \citet{tolstikhin2017wasserstein} for instance, proposed two different approaches for measuring the discrepancy between $\tilde{\mathcal{Z}}$ and $\mathcal{Z}$, namely the GAN-based and the {\it maximum mean discrepancy} (MMD)-based approaches. The GAN-based approach proposed in \cite{tolstikhin2017wasserstein} defines a discriminator network, $D_Z(p_Z,q_Z)$, to classify $\tilde{z}_j$s and $z_n$s as coming from `true' and `fake' distributions correspondingly and proposes a min-max adversarial optimization for learning $\phi$ and $D_Z$. This approach could be thought as a Fenchel conjugate of some $f$-divergence between $p_Z$ and $q_Z$. The MMD-based approach, on the other hand, utilizes a positive-definite reproducing kernel $k:Z\times Z\rightarrow \mathbb{R}$ to measure the discrepancy between $\tilde{\mathcal{Z}}$ and $\mathcal{Z}$, however, the choice of the kernel remain a data-dependent design parameter.

An interesting alternative approach is to use the Wasserstein distance between $p_Z$ and $q_Z$. The reason being that Wasserstein metrics have been shown to be particularly beneficial for measuring the distance between distributions supported on non-overlapping low-dimensional manifolds. 
Following the work of \citet{arjovsky2017wasserstein}, this can be accomplished utilizing the Kantorovich-Rubinstein duality and through introducing a min-max problem, which leads to yet another adversarial training scheme similar the GAN-based method in \cite{tolstikhin2017wasserstein}. Note that, since elements of $\tilde{\mathcal{Z}}$ and $\mathcal{Z}$ are not paired an approach similar to Eq. \eqref{eq:wcsimple} could not be used to calculate the Wasserstein distance. In this paper, we propose to use the sliced-Wasserstein metric, \cite{rabin2011awasserstein,rabin2011bwasserstein,bonneel2015sliced,kolouri2016sliced,kolouri2017sliced,carriere2017sliced}, to measure the discrepancy between $p_Z$ and $q_Z$. We show that using the sliced-Wasserstein distance ameliorates the need for training an adversary network, and provides an efficient but yet simple numerical implementation. 

Before explaining our proposed approach, it is worthwhile to point out the benefits of learning autoencoders as generative models over GANs. In GANs, one needs to minimize a distance between $\{\psi(\tilde{z}_j)|\tilde{z}_j\sim q_Z\}_{j=1}^M$ and $\{x_n\}_{n=1}^M$ which are high-dimensional point clouds for which there are no correspondences between $\psi(\tilde{z}_j)$s and $x_n$s. 
For the autoencoders, on the other hand, there exists correspondences between the high-dimensional point clouds $\{x_n\}_{n=1}^M$ and $\{y_n=\psi(\phi(x_n))\}_{n=1}^M$, and the problem simplifies to matching the lower-dimensional point clouds $\{\phi(x_n)\}_{n=1}^M$ and $\{\tilde{z}_j\sim q_Z\}_{j=1}^M$. In other words, the encoder performs a nonlinear dimensionality reduction, that enables us to solve a much simpler problem compared to GANs. Next we introduce the details of our approach.

\section{Proposed method}

In what follows we first provide a brief review of the necessary equations to understand the Wasserstein and sliced-Wasserstein distances and then present our Sliced Wassersten Autoencoders (SWAE).

\subsection{Wasserstein distances}

The Wasserstein distance between probability measures $\rho_X$ and $\rho_Y$, with corresponding densities $d\rho_X=p_X(x)dx$ and $d\rho_Y= p_Y(y)dy$ is defined as:
\begin{equation}
W_c(p_X,p_Y)=inf_{\gamma\in \Gamma(\rho_X,\rho_Y)} \int_{X\times Y} c(x,y)d\gamma(x,y)
\label{eq:kantorovich}
\end{equation}
where $\Gamma(\rho_X,\rho_Y)$ is the set of all transportation plans (i.e. joint measures) with marginal densities $p_X$ and $p_Y$, and $c:X\times Y\rightarrow \mathbb{R}^+$ is the transportation cost. Eq. \eqref{eq:kantorovich} is known as the Kantorovich formulation of the optimal mass transportation problem, which seeks the optimal transportation plan between $p_X$ and $p_Y$. If there exist diffeomorphic mappings, $f:X\rightarrow Y$ (i.e.  transport maps) such that $y=f(x)$ and consequently, 
\begin{equation}
p_Y(y)=\int_X p_X(x)\delta(y-f(x))dx \xrightarrow[\text{a diffeomorphism}]{\text{When $f$ is}}~ p_Y(y)=det(Df^{-1}(y))p_X(f^{-1}(y))
\label{eq:mp}
\end{equation}
where $det(D\cdot)$ is the determinant of the Jacobian, then the Wasserstein distance could be defined based on the Monge formulation of the problem (see \cite{villani2008optimal} and \cite{kolouri2017optimal}) as:
\begin{equation}
W_c(p_X,p_Y)=min_{f\in MP} \int_{X} c(x,f(x))d\rho_X(x) 
\label{eq:monge}
\end{equation}
where $MP$ is the set of all diffeomorphisms that satisfy Eq. \eqref{eq:mp}. As can be seen from Eqs. \eqref{eq:kantorovich} and \eqref{eq:monge}, obtaining the Wasserstein distance requires solving an optimization problem. Various efficient optimization techniques have been proposed in the past (e.g.  \cite{cuturi2013sinkhorn,solomon2015convolutional,oberman2015efficient}). 

The case of one dimensional probability densities, $p_X$ and $p_Y$, is specifically interesting as the Wasserstein distance has a closed-form solution. Let $P_X$ and $P_Y$ be the cumulative distributions of one-dimensional probability distributions $p_X$ and $p_Y$, correspondingly. The Wassertein distance can then be calculated as:
\begin{equation}
W_c(p_X,p_Y)= \int_{0}^1 c(P_X^{-1}(\tau),P_Y^{-1}(\tau))d\tau
\label{eq:oneD}
\end{equation}
The closed-form solution of Wasserstein distance for one-dimensional probability densities motivates the definition of sliced-Wasserstein distances.

\begin{figure}
\centering
\includegraphics[width=1.\linewidth]{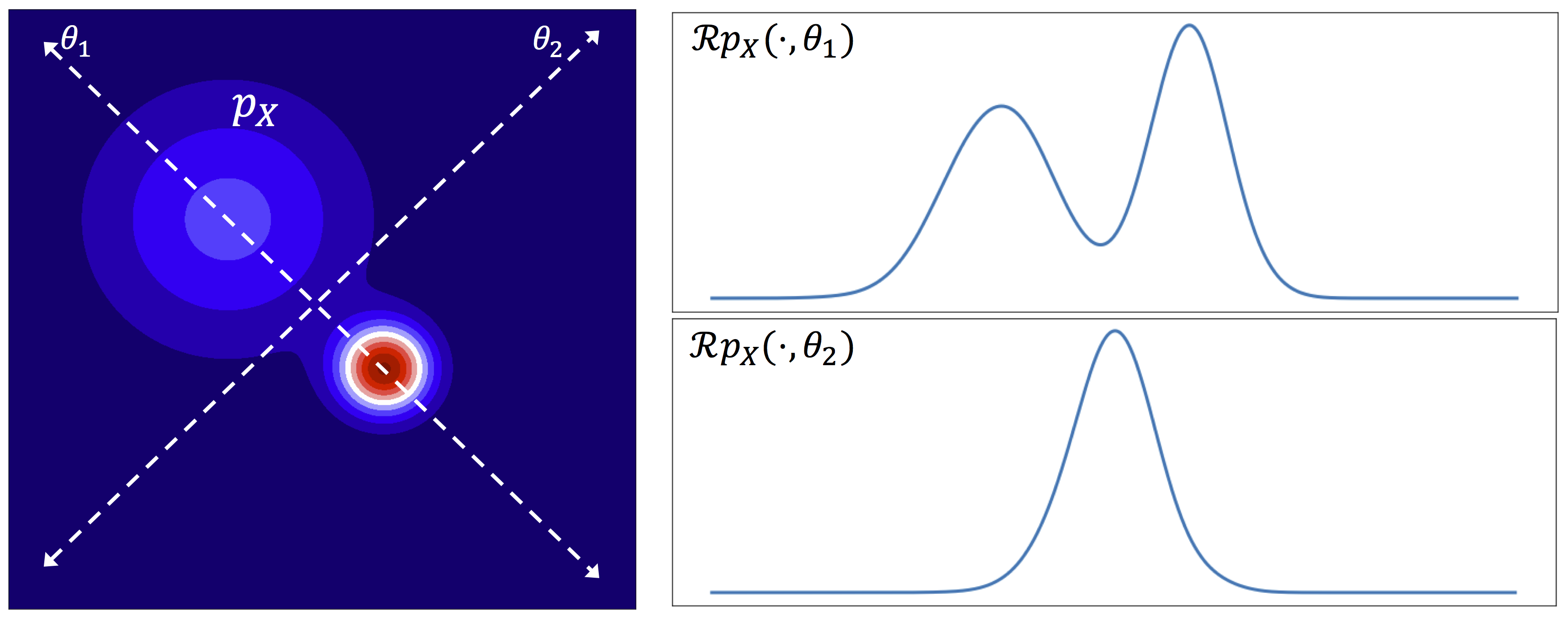}
\caption{Visualization of the slicing process defined in Eq. \eqref{eq:radon}}
\label{fig:slicing}
\end{figure}
\subsection{Sliced-Wasserstein distances}
The interest in the sliced-Wasserstein distance is due to the fact that it has very similar qualitative properties as the Wasserstein distance, but it is much easier to compute, since it only depends on one-dimensional computations. The sliced-Wasserstein distance was used in \cite{rabin2011awasserstein,rabin2011bwasserstein} to calculate barycenter of distributions and point clouds. \citet{bonneel2015sliced} provided a  nice theoretical overview of barycenteric calculations using the sliced-Wasserstein distance.  \citet{kolouri2016sliced} used this distance to define positive definite kernels for distributions and \citet{carriere2017sliced} used it as a distance for persistence diagrams. Sliced-Wasserstein was also recently  used  for learning Gaussian mixture models \cite{kolouri2017sliced}.  

The main idea behind the sliced-Wasserstein distance is to slice (i.e. project) higher-dimensional probability densities into sets of one-dimensional distributions and compare their one-dimensional representations via Wasserstein distance. The slicing/projection process is related to the field of Integral Geometry and specifically the Radon transform \cite{helgason2011radon}. The relevant result to our discussion is that a d-dimensional probability density $p_X$ could be uniquely represented as the set of its one-dimensional marginal distributions following the Radon transform and the Fourier slice theorem \cite{helgason2011radon}. These one dimensional marginal distributions of $p_X$ are defined as:
\begin{equation}
\mathcal{R}p_X(t;\theta)=\int_X p_X(x)\delta(t-\theta\cdot x)dx,~~\forall\theta\in\mathbb{S}^{d-1},~\forall t\in\mathbb{R}
\label{eq:radon}
\end{equation}
where $\mathbb{S}^{d-1}$ is the d-dimensional unit sphere. Note that for any fixed $\theta\in\mathbb{S}^{d-1}$, $\mathcal{R}p_X(\cdot;\theta)$ is a one-dimensional slice of distribution $p_X$. In other words, $\mathcal{R}p_X(\cdot;\theta)$ is a marginal distribution of $p_X$ that is obtained from integrating $p_X$ over the hyperplane orthogonal to $\theta$ (See Figure \ref{fig:slicing}). 
Utilizing the one-dimensional marginal distributions in Eq. \eqref{eq:radon}, the sliced Wasserstein distance could be defined as:
\begin{equation}
SW_c(p_X,p_Y)= \int_{\mathbb{S}^{d-1}} W_c(\mathcal{R} p_X(\cdot;\theta),\mathcal{R} p_Y(\cdot;\theta))d\theta
\label{eq:sw}
\end{equation}
Given that $\mathcal{R}p_X(\cdot;\theta)$ and $\mathcal{R}p_Y(\cdot;\theta)$ are one-dimensional the Wasserstein distance in the integrand has a closed-form solution as demonstrated in \eqref{eq:oneD}. The fact that $SW_c$ is a distance comes from $W_c$ being a distance. Moreover,
the two distances also induce the same topology, at least on compact sets \cite{santambrogio2015optimal}.

A natural transportation cost that has extensively studied in the past is the $\ell^2_2$, $c(x,y)=\|x-y\|^2_2$, for which there are theoretical guarantees on existence and uniqueness of transportation plans and maps (see \cite{santambrogio2015optimal} and \cite{villani2008optimal}). When $c(x,y)=\|x-y\|^2_2$ the following inequality bounds hold for the SW distance:
\begin{eqnarray}
SW_2(p_X,p_Y)\leq W_2(p_X,p_Y) \leq \alpha SW^\beta_2(p_X,p_Y)
\label{eq:inequalities}
\end{eqnarray}
where $\alpha$ is a constant. Chapter 5 in \cite{bonnotte2013unidimensional} proves this inequality with $\beta=(2(d+1))^{-1}$ (See \cite{santambrogio2015optimal} for more details). The inequalities in \eqref{eq:inequalities} is the main reason we can use the sliced Wasserstein distance, $SW_2$, as an approximation for $W_2$.

\subsection{Sliced-Wasserstein auto-encoder}
Our proposed formulation for the SWAE is as follows:
\begin{eqnarray}
\operatorname{argmin}_{\phi,\psi} W_c(p_X,p_Y)+\lambda SW_c(p_Z,q_Z)
\label{eq:swae}
\end{eqnarray}
where $\phi$ is the encoder, $\psi$ is the decoder, $p_X$ is the data distribution, $p_Y$ is the data distribution after encoding and decoding (Eq. \eqref{eq:py}), $p_Z$ is the distribution of the encoded data (Eq. \eqref{eq:rvt}), $q_Z$ is the predefined distribution (or a distribution we know how to sample from), and $\lambda$ is a hyperparameter that identifies the relative importance of the loss functions.  

To further clarify why we use the Wasserstein distance to measure the difference between $p_X$ and $p_Y$, but the \emph{sliced-Wasserstein} distance to measure the difference between $p_Z$ and $q_Z$, we reiterate that the Wasserstein distance for the first term can be solved via Eq. \eqref{eq:wcsimple} due to the existence of correspondences between $y_n$ and $x_n$ (i.e., we desire $x_n=y_n$), however, for $p_Z$ and $q_Z$, analogous correspondences between the $\tilde{z}_i$s and $z_j$s do not exist and therefore calculation of the Wasserstein distance requires an additional optimization step (e.g., in the form of an adversarial network). To avoid this additional optimization, while maintaining the favorable characteristics of the Wasserstein distance, we use the sliced-Wasserstein distance to measure the discrepancy between $p_Z$ and $q_Z$.

\section{Numerical optimization}

\begin{figure}[t]
\centering
\includegraphics[width=\linewidth]{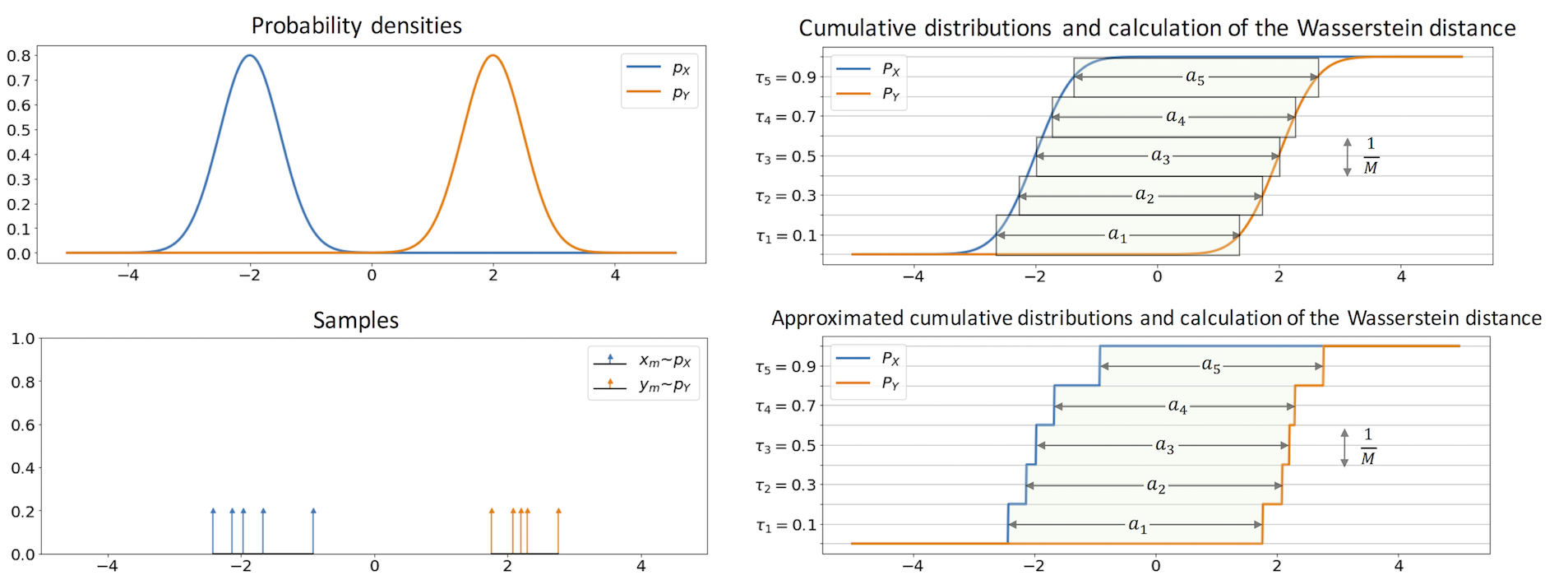}
\caption{The Wasserstein distance for one-dimensional probability distributions $p_X$ and $p_Y$ (top left) is calculated based on Eq. \eqref{eq:oneD}. For a numerical implementation, the integral in Eq. \eqref{eq:oneD} is substituted with $\frac{1}{M}\sum_{m=1}^{M}a_m$ where, $a_m=c(P^{-1}_X(\tau_m),P^{-1}_Y(\tau_m))$ (top right). When only samples from the distributions are available $x_n\sim p_X$ and $y_n\sim Y$ (bottom left), the Wasserstein distance is approximated by sorting $x_m$s and $y_m$s and letting $a_m=c(x_{i[m]},y_{j[m]})$, where $i[m]$ and $j[m]$ are the sorted indices (bottom right).}
\label{fig:numerics}
\end{figure}

\subsection{Numerical implementation of the Wasserstein distance in 1D}
The Wasserstein distance between two one-dimensional distributions $p_X$ and $p_Y$ is obtained from Eq. \eqref{eq:oneD}. The integral in Eq. \eqref{eq:oneD} could be numerically calculated using $\frac{1}{M}\sum_{m=1}^{M} a_m$, where $a_m=c(P_X^{-1}(\tau_m),P_Y^{-1}(\tau_m))$ and $\tau_m=\frac{2m-1}{2M}$ (see  Fig. \ref{fig:numerics}). In scenarios where only samples from the distributions are available, $x_m\sim p_X$ and $y_m\sim p_Y$, the empirical distributions can be estimated as $p_X=\frac{1}{M} \sum_{m=1}^M \delta_{x_m}$ and $p_Y=\frac{1}{M} \sum_{m=1}^M \delta_{y_m}$, where $\delta_{x_m}$ is the Dirac delta function centered at $x_m$. Therefore the corresponding empirical cumulative distribution of $p_X$ is $P_X(t)=\frac{1}{M}\sum_{m=1}^M u(t-x_m)$ where $u(.)$ is the step function ($P_Y$ is defined similarly). Sorting $x_m$s in an ascending order, such that $x_{i[m]}\leq x_{i[m+1]}$ and where $i[m]$ is the index of the sorted $x_m$s, it is straightforward to confirm that $P_X^{-1}(\tau_m)=x_{i[m]}$ (see Fig. \ref{fig:numerics} for a visual confirmation). Therefore, the Wasserstein distance can be approximated by first sorting $x_m$s and $y_m$s and then calculating:
\begin{equation}
W_c(p_X,p_Y)=\frac{1}{M}\sum_{m=1}^M c(x_{i[m]},y_{j[m]})
\label{eq:oneDnumerics}
\end{equation}
Eq. \eqref{eq:oneDnumerics} turns the problem of calculating the Wasserstein distance for two one-dimensional probability densities from their samples into a sorting problem that can be solved efficiently $(\mathcal{O}(M)$ best case and $\mathcal{O}(M log(M))$ worst case). 

\subsection{Slicing empirical distributions} 
In scenarios where only samples from the d-dimensional distribution, $p_X$, are available, $x_m\sim p_X$, the empirical distribution can be estimated as $p_X=\frac{1}{M} \sum_{m=1}^M \delta_{x_m}$. Following Eq. \eqref{eq:radon} it is straightforward to show that the marginal distributions (i.e. slices) of the empirical distribution, $p_X$, are obtained from:
\begin{equation}
\mathcal{R}p_X(t,\theta)= \frac{1}{M} \sum_{m=1}^M \delta(t-x_m\cdot\theta),~ \forall\theta\in \mathbb{S}^{d-1}, \text{and }~\forall t\in \mathbb{R}
\label{eq:radonnumerics}
\end{equation}
see the supplementary material for a proof.

\subsection{Minimizing sliced-Wasserstein via random slicing}

Minimizing the sliced-Wasserstein distance (i.e. as in the second term of Eq. \ref{eq:swae}) requires an integration over the unit sphere in $\mathbb{R}^{d}$, i.e., $\mathbb{S}^{d-1}$. In practice, this integration is substituted by a summation over a finite set $\Theta\subset \mathbb{S}^{d-1}$,
\begin{equation*}
\operatorname{min}_{\phi} SW_c(p_Z,q_Z)\approx \operatorname{min}_{\phi}\frac{1}{|\Theta|}\sum_{\theta_l\in\Theta} W_c(\mathcal{R}p_Z(\cdot;\theta_l),\mathcal{R}q_Z(\cdot;\theta_l))
\end{equation*}
Note that $SW_c(p_Z,q_Z)=\mathbb{E}_{\mathbb{S}^{(d-1)}}(W_c(\mathcal{R}p_Z(\cdot;\theta),\mathcal{R}q_Z(\cdot;\theta)))$. Moreover, the global minimum for $SW_c(p_Z,q_Z)$ is also a global minimum for each $W_c(\mathcal{R}p_Z(\cdot;\theta_l),\mathcal{R}q_Z(\cdot;\theta_l))$. A fine sampling of $\mathbb{S}^{d-1}$ is required for  a good approximation of $SW_c(p_Z,q_Z)$. Such sampling, however,  becomes prohibitively expensive as the dimension of the embedding space grows. Alternatively, following the approach presented by \citet{rabin2011awasserstein}, and later by \citet{bonneel2015sliced} and subsequently by \citet{kolouri2017sliced}, we utilize random samples of $\mathbb{S}^{d-1}$ at each minimization step to approximate the sliced-Wasserstein distance. Intuitively, if $p_Z$ and $q_Z$ are similar, then their projections with respect to any finite subset of $\mathbb{S}^{d-1}$ would also be similar. This leads to a stochastic gradient descent scheme where in addition to the random sampling of the input data, we also random sample the projection angles from $\mathbb{S}^{d-1}$.  


\subsection{Putting it all together}
To optimize the proposed SWAE objective function in Eq. \eqref{eq:swae} we use a stochastic gradient descent scheme as described here. In each iteration, let $\{x_m\sim p_X\}_{m=1}^M$ and $\{\tilde{z}_m\sim q_Z\}_{m=1}^M$  be i.i.d random samples from the input data and the predefined distribution, $q_Z$, correspondingly. Let $\{\theta_l\}_{l=1}^L$ be randomly sampled from a uniform distribution on $\mathbb{S}^{d-1}$. Then using the numerical approximations described in this section, the loss function in Eq. \eqref{eq:swae} can be rewritten as:
\begin{equation}
\mathcal{L}(\phi,\psi)=\frac{1}{M}\sum_{m=1}^M c(x_m,\psi(\phi(x_m)))+\frac{\lambda}{LM}\sum_{l=1}^L\sum_{m=1}^M c(\theta_l\cdot\tilde{z}_{i[m]},\theta_l\cdot \phi(x_{j[m]}))
\label{eq:swaenumerics}
\end{equation}
where $i[m]$ and $j[m]$ are the indices of sorted $\theta_l\cdot\tilde{z}_m$s and $\theta_l\cdot\phi(x_m)$ with respect to $m$, correspondingly. The steps of our proposed method are presented in Algorithm \ref{alg:algo1}. It is worth pointing out that sorting is by itself an optimization problem (which can be solved very efficiently), and therefore the sorting followed by the gradient descent update on $\phi$ and $\psi$ is in essence a min-max problem, which is being solved in an alternating fashion. 

\begin{algorithm}[t!]
\caption{\ Sliced-Wasserstein Auto-Encoder (SWAE)}
\label{alg:algo1}
\begin{algorithmic}
\REQUIRE Regularization coefficient $\lambda$, and number of random projections, $L$. 
\\Initialize the parameters of the encoder, $\phi$, and decoder, $\psi$
\WHILE{$\phi$ and $\psi$ have not converged}
\STATE Sample $\{x_1,...,x_M\}$ from training set (i.e. $p_X$)
\STATE Sample $\{\tilde{z}_1,...,\tilde{z}_M\}$ from $q_Z$
\STATE Sample $\{\theta_1,...,\theta_L\}$ from $\mathbb{S}^{K-1}$
\STATE Sort $\theta_l\cdot \tilde{z}_M$ such that $\theta_l\cdot \tilde{z}_{i[m]}\leq\theta_l\cdot \tilde{z}_{i[m+1]}$ 
\STATE Sort $\theta_l\cdot \phi(x_m)$ such that $\theta_l\cdot \phi(x_{j[m]})\leq\theta_l\cdot \phi(x_{j[m+1]})$ 
\STATE Update $\phi$ and $\psi$ by descending
\begin{equation*}
\sum_{m=1}^M c(x_m,\psi(\phi(x_m)))+\lambda\sum_{l=1}^L\sum_{m=1}^M c(\theta_l\cdot\tilde{z}_{i[m]},\theta_l\cdot \phi(x_{j[m]}))
\end{equation*}
\ENDWHILE
\end{algorithmic}
\end{algorithm}

\section{Experiments}

Here we show the results of SWAE for two mid-size image datasets, namely the MNIST dataset \cite{lecun1998mnist}, 
and the CelebFaces Attributes Dataset (CelebA) \cite{liu2015faceattributes}. For the encoder and the decoder we used mirrored classic deep convolutional neural networks with 2D average poolings and leaky rectified linear units (Leaky-ReLu) as the activation functions. The implementation details are included in the Supplementary material.

\begin{figure}[t]
\centering
\includegraphics[width=\linewidth]{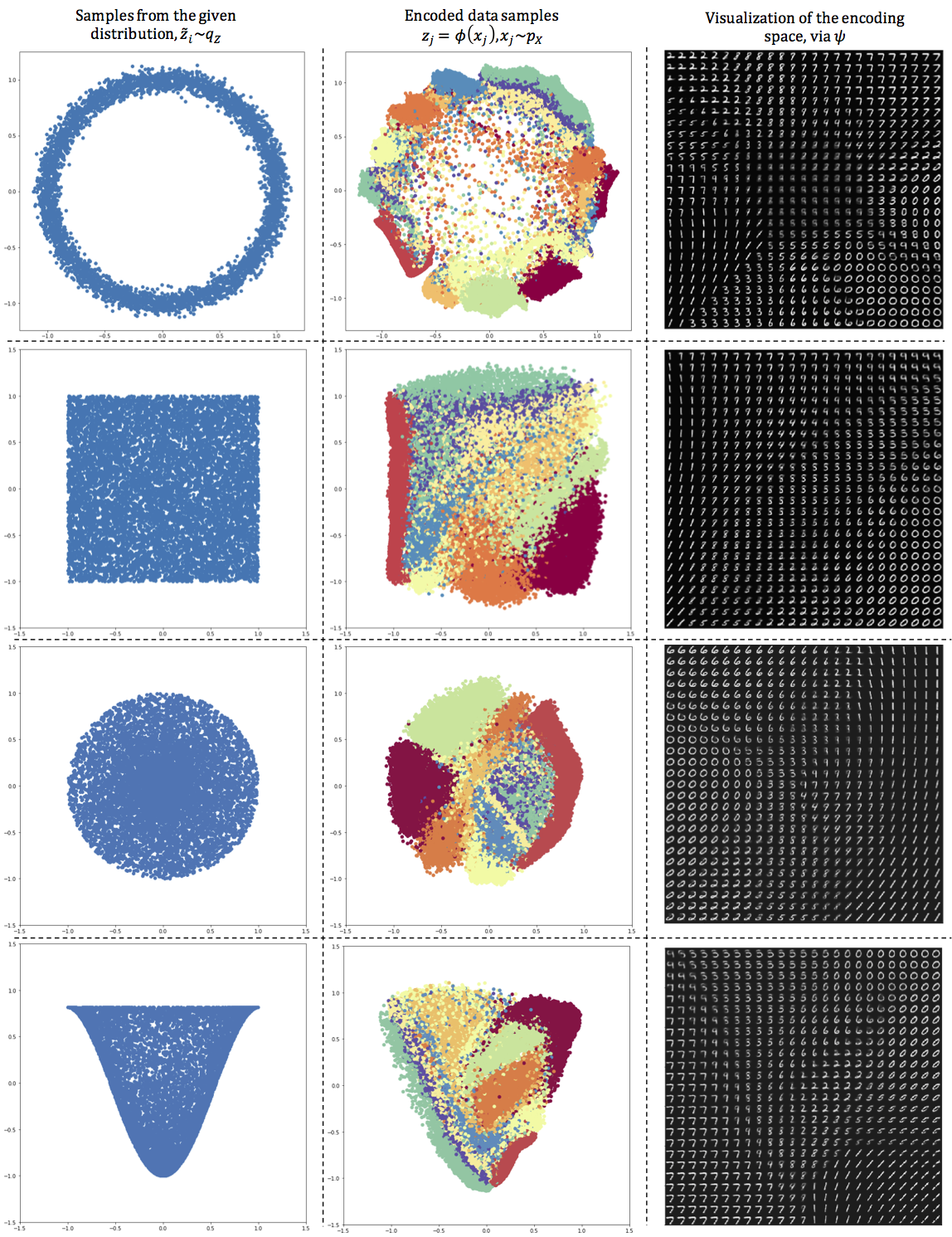}
\caption{The results of SWAE on the MNIST dataset for three different distributions as ,$q_Z$, namely the ring distribution, the uniform distribution, and the circle distribution. Note that the far right visualization is showing the decoding of a $25\times 25$ grid in $[-1,1]^2$ (in the encoding space).}
\label{fig:swae}
\end{figure}

For the MNIST dataset, we designed a deep convolutional encoder that embeds the handwritten digits into a two-dimensional embedding space (for visualization). To demonstrate the capability of SWAE on matching distributions $p_Z$ and $q_Z$ in the embedding/encoder space we chose four different $q_Z$s, namely the ring distribution, the uniform distribution, a circle distribution, and a bowl distribution. Figure \ref{fig:swae} shows the results of our experiment on the MNIST dataset. The left column shows samples from $q_Z$, the middle column shows $\phi(x_n)$s for the trained $\phi$ and the color represent the labels (note that the labels were only used for visualization). Finally, the right column depicts a $25\times 25$ grid in $[-1,1]^2$ through the trained decoder $\psi$. As can be seen, the embedding/encoder space closely follows the predefined $q_Z$, while the space remains decodable. The implementation details are included in the supplementary material. 

\begin{figure}[t]
\centering
\includegraphics[width=\linewidth]{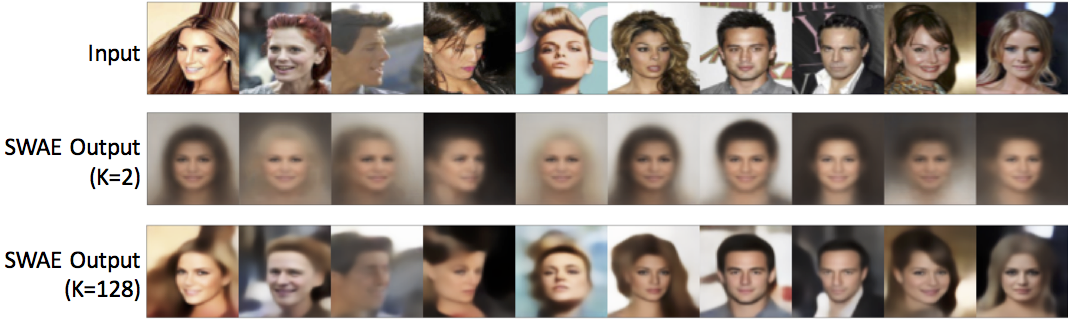}
\caption{Trained SWAE outputs for sample input images with different embedding spaces of size $K=2$ and $K=128$.}
\label{fig:celebAauto}
\end{figure}

The CelebA face dataset contains a higher degree of variations compared to the MNIST dataset and therefore a two-dimensional  embedding space does not suffice to capture the variations in this dataset. Therefore, while the SWAE loss function still goes down and the network achieves a good match between $p_Z$ and $q_Z$ the decoder is unable to match $p_X$ and $p_Y$. Therefore, a higher-dimensional embedding/encoder space is needed. In our experiments for this dataset we chose a $(K=128)-$dimensional embedding space. Figure \ref{fig:celebAauto} demonstrates the outputs of trained SWAEs with $K=2$ and $K=128$ for sample input images. The input images were resized to $64\times 64$ and then fed to our autoencoder structure. 

\begin{figure}[t]
\centering
\includegraphics[width=\linewidth]{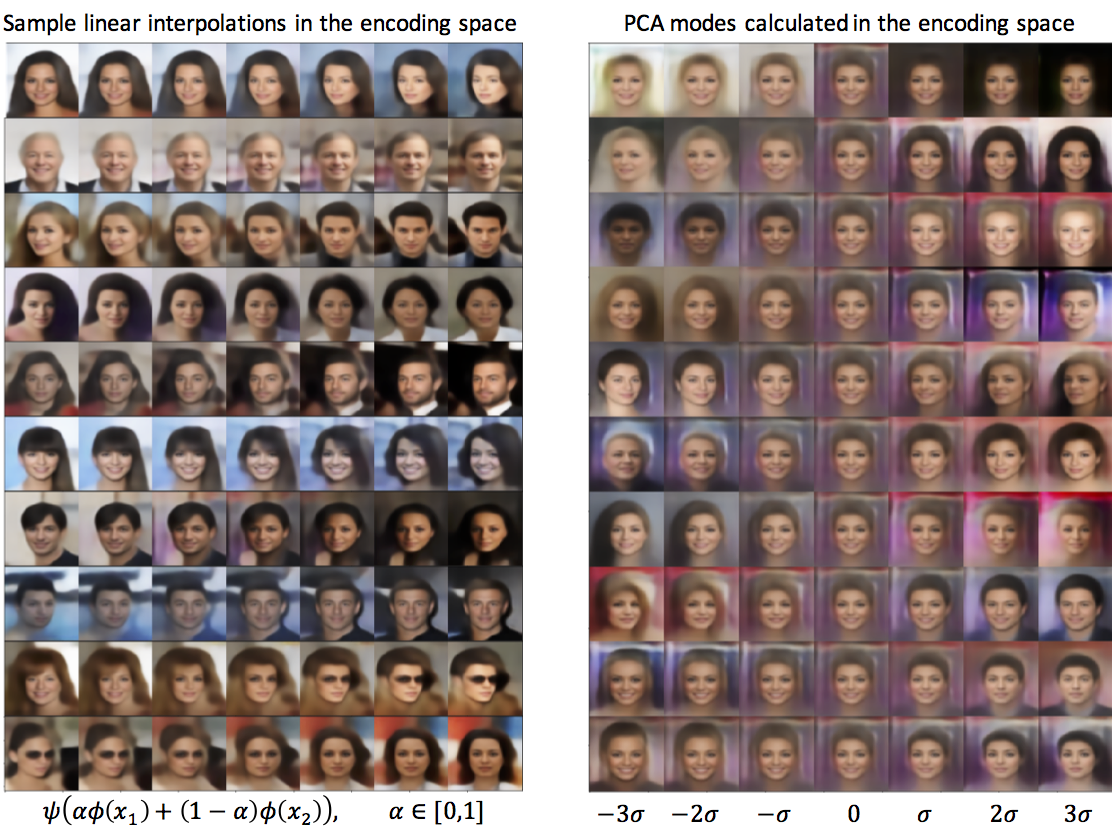}
\caption{The results of SWAE on the CelebA face  dataset with a 128-dimensional uniform distribution as $q_Z$. Linear interpolation in the encoding space for random samples (on the right) and the first 10 PCA components calculated in the encoding space.}
\label{fig:swae_faces}
\end{figure}

For CelebA dataset we set $q_Z$ to be a $(K=128)$-dimensional uniform distribution and trained our SWAE on the CelebA dataset. Given the convex nature of $q_Z$, any linear combination of the encoded faces should also result in a new face. Having that in mind, we ran two experiments in the embedding space to check that in fact the embedding space satisfies this convexity assumption. First we calculated linear interpolations of sampled pairs of faces in the embedding space and fed the interpolations to the decoder network to visualize the corresponding faces. Figure \ref{fig:swae_faces}, left column, shows the interpolation results for random pairs of encoded faces. It is clear that the interpolations remain faithful as expected from a uniform $q_Z$. Finally, we performed Principle Component Analysis (PCA) of the encoded faces and visualized the faces corresponding to these principle components via $\psi$. The PCA components are shown on the left column of Figure \ref{fig:swae_faces}. Various interesting modes including, hair color, skin color, gender, pose, etc. can be observed in the PC components. 

\section{Conclusions}

We introduced Sliced Wasserstein Autoencoders (SWAE), which enable one to shape the distribution of the encoded samples to any samplable distribution. We theoretically showed that utilizing the sliced Wasserstein distance as a dissimilarity measure between the distribution of the encoded samples and a predefined distribution ameliorates the need for training an adversarial network in the embedding space. In addition, we provided a simple and efficient numerical scheme for this problem, which only relies on few inner products and sorting operations in each SGD iteration. We further demonstrated the capability of our method on two mid-size image datasets, namely the MNIST dataset and the CelebA face dataset and showed results comparable to the techniques that rely on additional adversarial trainings. Our implementation is publicly available \footnote{\url{https://github.com/skolouri/swae}}.


\section*{Acknowledgments}

This work was partially supported by NSF (CCF 1421502). The authors would like to thank Drs. Dejan Slep\'{c}ev, and Heiko Hoffmann for their invaluable inputs and many hours of constructive conversations.

\bibliographystyle{plainnat}
\bibliography{swae}

\clearpage

\begin{figure}
\centering
\includegraphics[width=\linewidth]{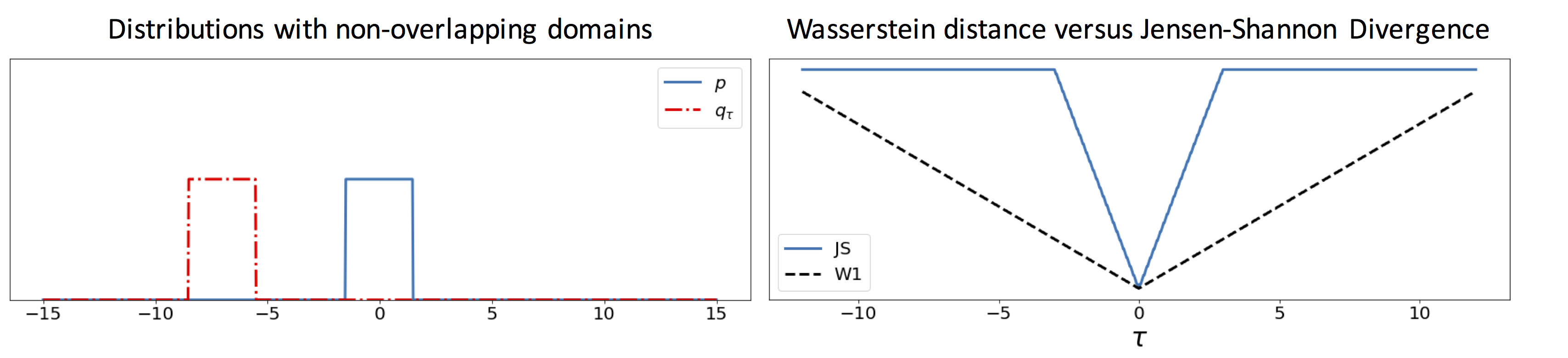}
\caption{These plots show $W_1(p,q_{\tau})$ and $JS(p,q_{\tau})$ where $p$ is a uniform distribution around zero and $q_{\tau}(x)=p(x-\tau)$. It is clear that JS divergence does not provide a usable gradient when distributions are supported on non-overlapping domains.}
\label{fig:energies}
\end{figure}

\section*{Supplementary Material}
\subsection*{Comparison of different distances}
Following the example by \citet{arjovsky2017wasserstein} and later \citet{kolouri2017sliced} here we show a simple example comparing the Jensen-Shannon divergence with the Wasserstein distance. First note that the Jensen-Shannon divergence is defined as, 
\begin{equation*}
JS(p,q)=KL(p,\frac{p+q}{2})+KL(q,\frac{p+q}{2})
\end{equation*}
where $KL(p,q)= \int_X p(x)log(\frac{p(x)}{q(x)})dx$ is the Kullback-Leibler divergence. Now consider the following densities, $p(x)$ be a uniform distribution around zero and let $q_\tau(x)=p(x-\tau)$ be a shifted version of the $p$. Figure \ref{fig:energies} show $W_1(p,q_\tau)$ and $JS(p,q_\tau)$ as a function of $\tau$. As can be seen the JS divergence fails to provide a useful gradient when the distributions are supported on non-overlapping domains.

\subsection*{Log-likelihood}
To maximize (minimize) the similarity (dissimilarity)  between $p_Z$ and $q_Z$, we can write :
\begin{eqnarray*}
\operatorname{argmax}_\phi \int_Z p_Z(z)log(q_Z(z))dz&=& \int_Z\int_X p_X(x)\delta(z-\phi(x))log(q_Z(z))dxdz\\
&=& \int_X p_X(x)log(q_Z(\phi(x)))dx
\end{eqnarray*}
where we replaced $p_Z$ with Eq. \eqref{eq:rvt}. Furthermore, it is straightforward to show:
\begin{eqnarray*}
\operatorname{argmax}_\phi \int_Z p_Z(z)log(q_Z(z))dz &=& \operatorname{argmax}_\phi \int_Z p_Z(z)log(\frac{q_Z(z)}{p_Z(z)})dz \\
&=&  \operatorname{argmin}_\phi D_{KL}(p_Z,q_Z)
\end{eqnarray*}
\subsection*{Slicing empirical distributions}

Here we calculate a Radon slice of the empirical distribution $p_X(x)=\frac{1}{M}\sum_{m=1}^M \delta(x-x_m)$ with respect to $\theta\in\mathbb{S}^{d-1}$. Using the definition of the Radon transform in Eq. \eqref{eq:radon} and RVT in Eq. \eqref{eq:rvt} we have:
\begin{eqnarray*}
\mathcal{R}p_X(t,\theta)&=&\int_X p_X(x)\delta(t-\theta\cdot x)dx \\
&=& \frac{1}{M}\sum_{m=1}^M\int_X \delta(x-x_m)\delta(t-\theta\cdot x)dx\\
&=& \frac{1}{M}\sum_{m=1}^M\delta(t-\theta\cdot x_m)
\end{eqnarray*}

\subsection*{Simple manifold learning experiment}

\begin{figure}[t]
\centering
\includegraphics[width=\linewidth]{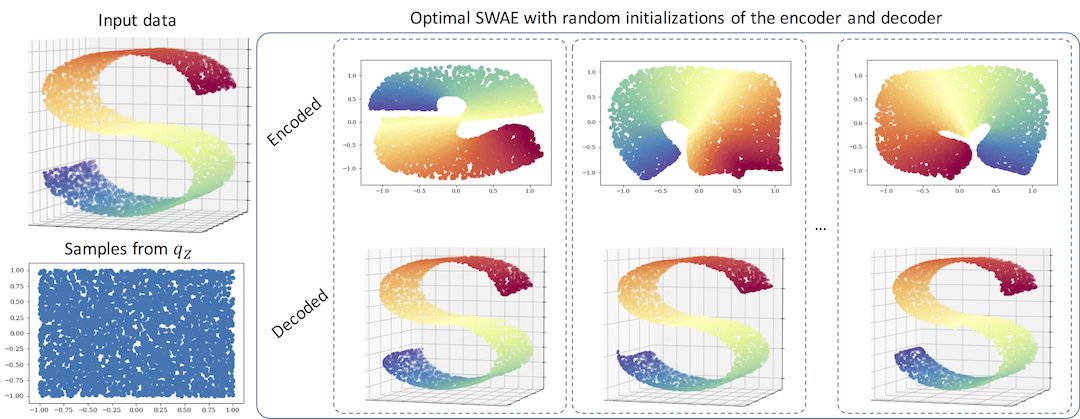}
\caption{Different runs of SWAE to embed a 3D nonlinear manifold into a 2D uniform distribution.}
\label{fig:swaeManifold3D}
\end{figure}

Figure \ref{fig:swaeManifold3D} demonstrates the results of SWAE with random initializations to embed a 2D manifold in $\mathbb{R}^3$ to a 2D uniform distribution. 

\input{iPythonNotebook.tex}

\end{document}

%% file: iPythonNotebookConfig.tex
    \usepackage[T1]{fontenc}
    \usepackage{mathpazo}

    \usepackage{graphicx}
    \makeatletter
    \def\maxwidth{\ifdim\Gin@nat@width>\linewidth\linewidth
    \else\Gin@nat@width\fi}
    \makeatother
    
    \usepackage{caption}

    \usepackage{adjustbox} 
    \usepackage{xcolor} 
    \usepackage{enumerate} 
    \usepackage{geometry} 
    \usepackage{amsmath} 
    \usepackage{amssymb} 
    \usepackage{textcomp} 
    \AtBeginDocument{%
        \def\PYZsq{\textquotesingle}
    }
    \usepackage{upquote} 
    \usepackage{eurosym} 
    \usepackage[mathletters]{ucs} 
    \usepackage[utf8x]{inputenc} 
    \usepackage{fancyvrb} 
    \usepackage{grffile} 
    \usepackage{hyperref}
    \usepackage{longtable} 
    \usepackage{booktabs}  
    \usepackage[inline]{enumitem} 
    \usepackage[normalem]{ulem} 

    \definecolor{urlcolor}{rgb}{0,.145,.698}
    \definecolor{linkcolor}{rgb}{.71,0.21,0.01}
    \definecolor{citecolor}{rgb}{.12,.54,.11}

    \definecolor{ansi-black}{HTML}{3E424D}
    \definecolor{ansi-black-intense}{HTML}{282C36}
    \definecolor{ansi-red}{HTML}{E75C58}
    \definecolor{ansi-red-intense}{HTML}{B22B31}
    \definecolor{ansi-green}{HTML}{00A250}
    \definecolor{ansi-green-intense}{HTML}{007427}
    \definecolor{ansi-yellow}{HTML}{DDB62B}
    \definecolor{ansi-yellow-intense}{HTML}{B27D12}
    \definecolor{ansi-blue}{HTML}{208FFB}
    \definecolor{ansi-blue-intense}{HTML}{0065CA}
    \definecolor{ansi-magenta}{HTML}{D160C4}
    \definecolor{ansi-magenta-intense}{HTML}{A03196}
    \definecolor{ansi-cyan}{HTML}{60C6C8}
    \definecolor{ansi-cyan-intense}{HTML}{258F8F}
    \definecolor{ansi-white}{HTML}{C5C1B4}
    \definecolor{ansi-white-intense}{HTML}{A1A6B2}

    \providecommand{\tightlist}{%
      \setlength{\itemsep}{0pt}\setlength{\parskip}{0pt}}
    \DefineVerbatimEnvironment{Highlighting}{Verbatim}{commandchars=\\\{\}}

\makeatletter
\def\PY@reset{\let\PY@it=\relax \let\PY@bf=\relax%
    \let\PY@ul=\relax \let\PY@tc=\relax%
    \let\PY@bc=\relax \let\PY@ff=\relax}
\def\PY@tok#1{\csname PY@tok@#1\endcsname}
\def\PY@toks#1+{\ifx\relax#1\empty\else%
    \PY@tok{#1}\expandafter\PY@toks\fi}
\def\PY@do#1{\PY@bc{\PY@tc{\PY@ul{%
    \PY@it{\PY@bf{\PY@ff{#1}}}}}}}
\def\PY#1#2{\PY@reset\PY@toks#1+\relax+\PY@do{#2}}

\expandafter\def\csname PY@tok@gd\endcsname{\def\PY@tc##1{\textcolor[rgb]{0.63,0.00,0.00}{##1}}}
\expandafter\def\csname PY@tok@gu\endcsname{\let\PY@bf=\textbf\def\PY@tc##1{\textcolor[rgb]{0.50,0.00,0.50}{##1}}}
\expandafter\def\csname PY@tok@gt\endcsname{\def\PY@tc##1{\textcolor[rgb]{0.00,0.27,0.87}{##1}}}
\expandafter\def\csname PY@tok@gs\endcsname{\let\PY@bf=\textbf}
\expandafter\def\csname PY@tok@gr\endcsname{\def\PY@tc##1{\textcolor[rgb]{1.00,0.00,0.00}{##1}}}
\expandafter\def\csname PY@tok@cm\endcsname{\let\PY@it=\textit\def\PY@tc##1{\textcolor[rgb]{0.25,0.50,0.50}{##1}}}
\expandafter\def\csname PY@tok@vg\endcsname{\def\PY@tc##1{\textcolor[rgb]{0.10,0.09,0.49}{##1}}}
\expandafter\def\csname PY@tok@vi\endcsname{\def\PY@tc##1{\textcolor[rgb]{0.10,0.09,0.49}{##1}}}
\expandafter\def\csname PY@tok@vm\endcsname{\def\PY@tc##1{\textcolor[rgb]{0.10,0.09,0.49}{##1}}}
\expandafter\def\csname PY@tok@mh\endcsname{\def\PY@tc##1{\textcolor[rgb]{0.40,0.40,0.40}{##1}}}
\expandafter\def\csname PY@tok@cs\endcsname{\let\PY@it=\textit\def\PY@tc##1{\textcolor[rgb]{0.25,0.50,0.50}{##1}}}
\expandafter\def\csname PY@tok@ge\endcsname{\let\PY@it=\textit}
\expandafter\def\csname PY@tok@vc\endcsname{\def\PY@tc##1{\textcolor[rgb]{0.10,0.09,0.49}{##1}}}
\expandafter\def\csname PY@tok@il\endcsname{\def\PY@tc##1{\textcolor[rgb]{0.40,0.40,0.40}{##1}}}
\expandafter\def\csname PY@tok@go\endcsname{\def\PY@tc##1{\textcolor[rgb]{0.53,0.53,0.53}{##1}}}
\expandafter\def\csname PY@tok@cp\endcsname{\def\PY@tc##1{\textcolor[rgb]{0.74,0.48,0.00}{##1}}}
\expandafter\def\csname PY@tok@gi\endcsname{\def\PY@tc##1{\textcolor[rgb]{0.00,0.63,0.00}{##1}}}
\expandafter\def\csname PY@tok@gh\endcsname{\let\PY@bf=\textbf\def\PY@tc##1{\textcolor[rgb]{0.00,0.00,0.50}{##1}}}
\expandafter\def\csname PY@tok@ni\endcsname{\let\PY@bf=\textbf\def\PY@tc##1{\textcolor[rgb]{0.60,0.60,0.60}{##1}}}
\expandafter\def\csname PY@tok@nl\endcsname{\def\PY@tc##1{\textcolor[rgb]{0.63,0.63,0.00}{##1}}}
\expandafter\def\csname PY@tok@nn\endcsname{\let\PY@bf=\textbf\def\PY@tc##1{\textcolor[rgb]{0.00,0.00,1.00}{##1}}}
\expandafter\def\csname PY@tok@no\endcsname{\def\PY@tc##1{\textcolor[rgb]{0.53,0.00,0.00}{##1}}}
\expandafter\def\csname PY@tok@na\endcsname{\def\PY@tc##1{\textcolor[rgb]{0.49,0.56,0.16}{##1}}}
\expandafter\def\csname PY@tok@nb\endcsname{\def\PY@tc##1{\textcolor[rgb]{0.00,0.50,0.00}{##1}}}
\expandafter\def\csname PY@tok@nc\endcsname{\let\PY@bf=\textbf\def\PY@tc##1{\textcolor[rgb]{0.00,0.00,1.00}{##1}}}
\expandafter\def\csname PY@tok@nd\endcsname{\def\PY@tc##1{\textcolor[rgb]{0.67,0.13,1.00}{##1}}}
\expandafter\def\csname PY@tok@ne\endcsname{\let\PY@bf=\textbf\def\PY@tc##1{\textcolor[rgb]{0.82,0.25,0.23}{##1}}}
\expandafter\def\csname PY@tok@nf\endcsname{\def\PY@tc##1{\textcolor[rgb]{0.00,0.00,1.00}{##1}}}
\expandafter\def\csname PY@tok@si\endcsname{\let\PY@bf=\textbf\def\PY@tc##1{\textcolor[rgb]{0.73,0.40,0.53}{##1}}}
\expandafter\def\csname PY@tok@s2\endcsname{\def\PY@tc##1{\textcolor[rgb]{0.73,0.13,0.13}{##1}}}
\expandafter\def\csname PY@tok@nt\endcsname{\let\PY@bf=\textbf\def\PY@tc##1{\textcolor[rgb]{0.00,0.50,0.00}{##1}}}
\expandafter\def\csname PY@tok@nv\endcsname{\def\PY@tc##1{\textcolor[rgb]{0.10,0.09,0.49}{##1}}}
\expandafter\def\csname PY@tok@s1\endcsname{\def\PY@tc##1{\textcolor[rgb]{0.73,0.13,0.13}{##1}}}
\expandafter\def\csname PY@tok@dl\endcsname{\def\PY@tc##1{\textcolor[rgb]{0.73,0.13,0.13}{##1}}}
\expandafter\def\csname PY@tok@ch\endcsname{\let\PY@it=\textit\def\PY@tc##1{\textcolor[rgb]{0.25,0.50,0.50}{##1}}}
\expandafter\def\csname PY@tok@m\endcsname{\def\PY@tc##1{\textcolor[rgb]{0.40,0.40,0.40}{##1}}}
\expandafter\def\csname PY@tok@gp\endcsname{\let\PY@bf=\textbf\def\PY@tc##1{\textcolor[rgb]{0.00,0.00,0.50}{##1}}}
\expandafter\def\csname PY@tok@sh\endcsname{\def\PY@tc##1{\textcolor[rgb]{0.73,0.13,0.13}{##1}}}
\expandafter\def\csname PY@tok@ow\endcsname{\let\PY@bf=\textbf\def\PY@tc##1{\textcolor[rgb]{0.67,0.13,1.00}{##1}}}
\expandafter\def\csname PY@tok@sx\endcsname{\def\PY@tc##1{\textcolor[rgb]{0.00,0.50,0.00}{##1}}}
\expandafter\def\csname PY@tok@bp\endcsname{\def\PY@tc##1{\textcolor[rgb]{0.00,0.50,0.00}{##1}}}
\expandafter\def\csname PY@tok@c1\endcsname{\let\PY@it=\textit\def\PY@tc##1{\textcolor[rgb]{0.25,0.50,0.50}{##1}}}
\expandafter\def\csname PY@tok@fm\endcsname{\def\PY@tc##1{\textcolor[rgb]{0.00,0.00,1.00}{##1}}}
\expandafter\def\csname PY@tok@o\endcsname{\def\PY@tc##1{\textcolor[rgb]{0.40,0.40,0.40}{##1}}}
\expandafter\def\csname PY@tok@kc\endcsname{\let\PY@bf=\textbf\def\PY@tc##1{\textcolor[rgb]{0.00,0.50,0.00}{##1}}}
\expandafter\def\csname PY@tok@c\endcsname{\let\PY@it=\textit\def\PY@tc##1{\textcolor[rgb]{0.25,0.50,0.50}{##1}}}
\expandafter\def\csname PY@tok@mf\endcsname{\def\PY@tc##1{\textcolor[rgb]{0.40,0.40,0.40}{##1}}}
\expandafter\def\csname PY@tok@err\endcsname{\def\PY@bc##1{\setlength{\fboxsep}{0pt}\fcolorbox[rgb]{1.00,0.00,0.00}{1,1,1}{\strut ##1}}}
\expandafter\def\csname PY@tok@mb\endcsname{\def\PY@tc##1{\textcolor[rgb]{0.40,0.40,0.40}{##1}}}
\expandafter\def\csname PY@tok@ss\endcsname{\def\PY@tc##1{\textcolor[rgb]{0.10,0.09,0.49}{##1}}}
\expandafter\def\csname PY@tok@sr\endcsname{\def\PY@tc##1{\textcolor[rgb]{0.73,0.40,0.53}{##1}}}
\expandafter\def\csname PY@tok@mo\endcsname{\def\PY@tc##1{\textcolor[rgb]{0.40,0.40,0.40}{##1}}}
\expandafter\def\csname PY@tok@kd\endcsname{\let\PY@bf=\textbf\def\PY@tc##1{\textcolor[rgb]{0.00,0.50,0.00}{##1}}}
\expandafter\def\csname PY@tok@mi\endcsname{\def\PY@tc##1{\textcolor[rgb]{0.40,0.40,0.40}{##1}}}
\expandafter\def\csname PY@tok@kn\endcsname{\let\PY@bf=\textbf\def\PY@tc##1{\textcolor[rgb]{0.00,0.50,0.00}{##1}}}
\expandafter\def\csname PY@tok@cpf\endcsname{\let\PY@it=\textit\def\PY@tc##1{\textcolor[rgb]{0.25,0.50,0.50}{##1}}}
\expandafter\def\csname PY@tok@kr\endcsname{\let\PY@bf=\textbf\def\PY@tc##1{\textcolor[rgb]{0.00,0.50,0.00}{##1}}}
\expandafter\def\csname PY@tok@s\endcsname{\def\PY@tc##1{\textcolor[rgb]{0.73,0.13,0.13}{##1}}}
\expandafter\def\csname PY@tok@kp\endcsname{\def\PY@tc##1{\textcolor[rgb]{0.00,0.50,0.00}{##1}}}
\expandafter\def\csname PY@tok@w\endcsname{\def\PY@tc##1{\textcolor[rgb]{0.73,0.73,0.73}{##1}}}
\expandafter\def\csname PY@tok@kt\endcsname{\def\PY@tc##1{\textcolor[rgb]{0.69,0.00,0.25}{##1}}}
\expandafter\def\csname PY@tok@sc\endcsname{\def\PY@tc##1{\textcolor[rgb]{0.73,0.13,0.13}{##1}}}
\expandafter\def\csname PY@tok@sb\endcsname{\def\PY@tc##1{\textcolor[rgb]{0.73,0.13,0.13}{##1}}}
\expandafter\def\csname PY@tok@sa\endcsname{\def\PY@tc##1{\textcolor[rgb]{0.73,0.13,0.13}{##1}}}
\expandafter\def\csname PY@tok@k\endcsname{\let\PY@bf=\textbf\def\PY@tc##1{\textcolor[rgb]{0.00,0.50,0.00}{##1}}}
\expandafter\def\csname PY@tok@se\endcsname{\let\PY@bf=\textbf\def\PY@tc##1{\textcolor[rgb]{0.73,0.40,0.13}{##1}}}
\expandafter\def\csname PY@tok@sd\endcsname{\let\PY@it=\textit\def\PY@tc##1{\textcolor[rgb]{0.73,0.13,0.13}{##1}}}

\def\PYZbs{\char`\\}
\def\PYZus{\char`\_}

\def\PYZsh{\char`\#}

\def\PYZdl{\char`\$}
\def\PYZhy{\char`\-}
\def\PYZsq{\char`\'}

\makeatother

    \definecolor{incolor}{rgb}{0.0, 0.0, 0.5}
    \definecolor{outcolor}{rgb}{0.545, 0.0, 0.0}

    \hypersetup{
      breaklinks=true,  
      colorlinks=true,
      urlcolor=urlcolor,
      linkcolor=linkcolor,
      citecolor=citecolor,
      }
    

%% file: iPythonNotebook.tex
\subsection*{\bf The implementation details of our algorithm}

\fontsize{10}{12}\selectfont
The following text walks you through the implementation of our Sliced Wasserstein Autoencoders (SWAE).

 To run this notebook you'll require the following packages:

 \begin{itemize}
 \tightlist
 \item
   Numpy
 \item
   Matplotlib
 \item
   tensorflow
 \item
   Keras
 \end{itemize}

     \begin{Verbatim}[commandchars=\\\{\}]
 {\color{incolor}In [{\color{incolor}1}]:} \PY{k+kn}{import} \PY{n+nn}{numpy} \PY{k+kn}{as} \PY{n+nn}{np}
         \PY{k+kn}{import} \PY{n+nn}{keras.utils}
         \PY{k+kn}{from} \PY{n+nn}{keras.layers} \PY{k+kn}{import} \PY{n}{Input}\PY{p}{,}\PY{n}{Dense}\PY{p}{,} \PY{n}{Flatten}\PY{p}{,}\PY{n}{Activation}
         \PY{k+kn}{from} \PY{n+nn}{keras.models} \PY{n+nn}{importload\PYZus{}model}\PY{p}{,}\PY{n}{Model}
         \PY{k+kn}{from} \PY{n+nn}{keras.layers} \PY{k+kn}{import} \PY{n}{Conv2D}\PY{p}{,} \PY{n}{UpSampling2D}\PY{p}{,} \PY{n}{AveragePooling2D}
         \PY{k+kn}{from} \PY{n+nn}{keras.layers} \PY{k+kn}{import} \PY{n}{LeakyReLU}\PY{p}{,}\PY{n}{Reshape}
         \PY{k+kn}{from} \PY{n+nn}{keras.preprocessing.image} \PY{k+kn}{import} \PY{n}{ImageDataGenerator}
         \PY{k+kn}{from} \PY{n+nn}{keras.optimizers} \PY{k+kn}{import} \PY{n}{RMSprop}
         \PY{k+kn}{from} \PY{n+nn}{keras.datasets} \PY{k+kn}{import} \PY{n}{mnist}
         \PY{k+kn}{from} \PY{n+nn}{keras.models} \PY{k+kn}{import} \PY{n}{save\PYZus{}model}
         \PY{k+kn}{from} \PY{n+nn}{keras} \PY{k+kn}{import} \PY{n}{backend} \PY{k}{as} \PY{n}{K}
         \PY{k+kn}{import} \PY{n+nn}{tensorflow} \PY{k+kn}{as} \PY{n+nn}{tf}
         \PY{k+kn}{import} \PY{n+nn}{matplotlib.pyplot} \PY{k+kn}{as} \PY{n+nn}{plt}
         \PY{k+kn}{from} \PY{n+nn}{IPython} \PY{k+kn}{import} \PY{n}{display}
         \PY{k+kn}{import} \PY{n+nn}{time}
 \end{Verbatim}

    \begin{Verbatim}[commandchars=\\\{\}]
Using TensorFlow backend.

    \end{Verbatim}

    \hypertarget{define-three-helper-function}{%
\subsection{Define three helper
function}\label{define-three-helper-function}}

\begin{itemize}
\tightlist
\item
  generateTheta(L,dim) -\textgreater{} Generates \(L\) random sampels
  from \(\mathbb{S}^{dim-1}\)
\item
  generateZ(batchsize,endim) -\textgreater{} Generates `batchsize'
  samples `endim' dimensional samples from \(q_Z\)
\item
  stitchImages(I,axis=0) -\textgreater{} Helps us with visualization
\end{itemize}

    \begin{Verbatim}[commandchars=\\\{\}]
{\color{incolor}In [{\color{incolor}2}]:} \PY{k}{def} \PY{n+nf}{generateTheta}\PY{p}{(}\PY{n}{L}\PY{p}{,}\PY{n}{endim}\PY{p}{)}\PY{p}{:}
            \PY{n}{theta\PYZus{}}\PY{o}{=}\PY{n}{np}\PY{o}{.}\PY{n}{random}\PY{o}{.}\PY{n}{normal}\PY{p}{(}\PY{n}{size}\PY{o}{=}\PY{p}{(}\PY{n}{L}\PY{p}{,}\PY{n}{endim}\PY{p}{)}\PY{p}{)}
            \PY{k}{for} \PY{n}{l} \PY{o+ow}{in} \PY{n+nb}{range}\PY{p}{(}\PY{n}{L}\PY{p}{)}\PY{p}{:}
                \PY{n}{theta\PYZus{}}\PY{p}{[}\PY{n}{l}\PY{p}{,}\PY{p}{:}\PY{p}{]}\PY{o}{=}\PY{n}{theta\PYZus{}}\PY{p}{[}\PY{n}{l}\PY{p}{,}\PY{p}{:}\PY{p}{]}\PY{o}{/}\PY{n}{np}\PY{o}{.}\PY{n}{sqrt}\PY{p}{(}\PY{n}{np}\PY{o}{.}\PY{n}{sum}\PY{p}{(}\PY{n}{theta\PYZus{}}\PY{p}{[}\PY{n}{l}\PY{p}{,}\PY{p}{:}\PY{p}{]}\PY{o}{*}\PY{o}{*}\PY{l+m+mi}{2}\PY{p}{)}\PY{p}{)}
            \PY{k}{return} \PY{n}{theta\PYZus{}}
        \PY{k}{def} \PY{n+nf}{generateZ}\PY{p}{(}\PY{n}{batchsize}\PY{p}{,}\PY{n}{endim}\PY{p}{)}\PY{p}{:}
            \PY{n}{z\PYZus{}}\PY{o}{=}\PY{l+m+mi}{2}\PY{o}{*}\PY{p}{(}\PY{n}{np}\PY{o}{.}\PY{n}{random}\PY{o}{.}\PY{n}{uniform}\PY{p}{(}\PY{n}{size}\PY{o}{=}\PY{p}{(}\PY{n}{batchsize}\PY{p}{,}\PY{n}{endim}\PY{p}{)}\PY{p}{)}\PY{o}{\PYZhy{}}\PY{l+m+mf}{0.5}\PY{p}{)}
            \PY{k}{return} \PY{n}{z\PYZus{}}
        \PY{k}{def} \PY{n+nf}{stitchImages}\PY{p}{(}\PY{n}{I}\PY{p}{,}\PY{n}{axis}\PY{o}{=}\PY{l+m+mi}{0}\PY{p}{)}\PY{p}{:}
            \PY{n}{n}\PY{p}{,}\PY{n}{N}\PY{p}{,}\PY{n}{M}\PY{p}{,}\PY{n}{K}\PY{o}{=}\PY{n}{I}\PY{o}{.}\PY{n}{shape}
            \PY{k}{if} \PY{n}{axis}\PY{o}{==}\PY{l+m+mi}{0}\PY{p}{:}
                \PY{n}{img}\PY{o}{=}\PY{n}{np}\PY{o}{.}\PY{n}{zeros}\PY{p}{(}\PY{p}{(}\PY{n}{N}\PY{o}{*}\PY{n}{n}\PY{p}{,}\PY{n}{M}\PY{p}{,}\PY{n}{K}\PY{p}{)}\PY{p}{)}
                \PY{k}{for} \PY{n}{i} \PY{o+ow}{in} \PY{n+nb}{range}\PY{p}{(}\PY{n}{n}\PY{p}{)}\PY{p}{:}
                    \PY{n}{img}\PY{p}{[}\PY{n}{i}\PY{o}{*}\PY{n}{N}\PY{p}{:}\PY{p}{(}\PY{n}{i}\PY{o}{+}\PY{l+m+mi}{1}\PY{p}{)}\PY{o}{*}\PY{n}{N}\PY{p}{,}\PY{p}{:}\PY{p}{,}\PY{p}{:}\PY{p}{]}\PY{o}{=}\PY{n}{I}\PY{p}{[}\PY{n}{i}\PY{p}{,}\PY{p}{:}\PY{p}{,}\PY{p}{:}\PY{p}{,}\PY{p}{:}\PY{p}{]}
            \PY{k}{else}\PY{p}{:}
                \PY{n}{img}\PY{o}{=}\PY{n}{np}\PY{o}{.}\PY{n}{zeros}\PY{p}{(}\PY{p}{(}\PY{n}{N}\PY{p}{,}\PY{n}{M}\PY{o}{*}\PY{n}{n}\PY{p}{,}\PY{n}{K}\PY{p}{)}\PY{p}{)}
                \PY{k}{for} \PY{n}{i} \PY{o+ow}{in} \PY{n+nb}{range}\PY{p}{(}\PY{n}{n}\PY{p}{)}\PY{p}{:}
                    \PY{n}{img}\PY{p}{[}\PY{p}{:}\PY{p}{,}\PY{n}{i}\PY{o}{*}\PY{n}{M}\PY{p}{:}\PY{p}{(}\PY{n}{i}\PY{o}{+}\PY{l+m+mi}{1}\PY{p}{)}\PY{o}{*}\PY{n}{M}\PY{p}{,}\PY{p}{:}\PY{p}{]}\PY{o}{=}\PY{n}{I}\PY{p}{[}\PY{n}{i}\PY{p}{,}\PY{p}{:}\PY{p}{,}\PY{p}{:}\PY{p}{,}\PY{p}{:}\PY{p}{]}
            \PY{k}{return} \PY{n}{img}
\end{Verbatim}

    \hypertarget{defining-the-encoderdecoder-as-keras-graphs}{%
\subsection{Defining the Encoder/Decoder as Keras
graphs}\label{defining-the-encoderdecoder-as-keras-graphs}}

    \begin{Verbatim}[commandchars=\\\{\}]
{\color{incolor}In [{\color{incolor}3}]:} \PY{n}{img}\PY{o}{=}\PY{n}{Input}\PY{p}{(}\PY{p}{(}\PY{l+m+mi}{28}\PY{p}{,}\PY{l+m+mi}{28}\PY{p}{,}\PY{l+m+mi}{1}\PY{p}{)}\PY{p}{)} \PY{c+c1}{\PYZsh{}Input image }
        \PY{n}{interdim}\PY{o}{=}\PY{l+m+mi}{128} \PY{c+c1}{\PYZsh{} This is the dimension of intermediate latent  }
                     \PY{c+c1}{\PYZsh{}(variable after convolution and before embedding)}
        \PY{n}{endim}\PY{o}{=}\PY{l+m+mi}{2} \PY{c+c1}{\PYZsh{} Dimension of the embedding space}
        \PY{n}{embedd}\PY{o}{=}\PY{n}{Input}\PY{p}{(}\PY{p}{(}\PY{n}{endim}\PY{p}{,}\PY{p}{)}\PY{p}{)} \PY{c+c1}{\PYZsh{}Keras input to Decoder}
        \PY{n}{depth}\PY{o}{=}\PY{l+m+mi}{16} \PY{c+c1}{\PYZsh{} This is a design parameter.}
        \PY{n}{L}\PY{o}{=}\PY{l+m+mi}{50} \PY{c+c1}{\PYZsh{} Number of random projections}
        \PY{n}{batchsize}\PY{o}{=}\PY{l+m+mi}{500}
\end{Verbatim}

    \hypertarget{define-encoder}{%
\subsubsection{Define Encoder}\label{define-encoder}}

    \begin{Verbatim}[commandchars=\\\{\}]
{\color{incolor}In [{\color{incolor}4}]:} \PY{n}{x}\PY{o}{=}\PY{n}{Conv2D}\PY{p}{(}\PY{n}{depth}\PY{o}{*}\PY{l+m+mi}{1}\PY{p}{,} \PY{p}{(}\PY{l+m+mi}{3}\PY{p}{,} \PY{l+m+mi}{3}\PY{p}{)}\PY{p}{,} \PY{n}{padding}\PY{o}{=}\PY{l+s+s1}{\PYZsq{}}\PY{l+s+s1}{same}\PY{l+s+s1}{\PYZsq{}}\PY{p}{)}\PY{p}{(}\PY{n}{img}\PY{p}{)}
        \PY{n}{x}\PY{o}{=}\PY{n}{LeakyReLU}\PY{p}{(}\PY{n}{alpha}\PY{o}{=}\PY{l+m+mf}{0.2}\PY{p}{)}\PY{p}{(}\PY{n}{x}\PY{p}{)}
        \PY{c+c1}{\PYZsh{} x=BatchNormalization(momentum=0.8)(x)}
        \PY{n}{x}\PY{o}{=}\PY{n}{Conv2D}\PY{p}{(}\PY{n}{depth}\PY{o}{*}\PY{l+m+mi}{1}\PY{p}{,} \PY{p}{(}\PY{l+m+mi}{3}\PY{p}{,} \PY{l+m+mi}{3}\PY{p}{)}\PY{p}{,} \PY{n}{padding}\PY{o}{=}\PY{l+s+s1}{\PYZsq{}}\PY{l+s+s1}{same}\PY{l+s+s1}{\PYZsq{}}\PY{p}{)}\PY{p}{(}\PY{n}{x}\PY{p}{)}
        \PY{n}{x}\PY{o}{=}\PY{n}{LeakyReLU}\PY{p}{(}\PY{n}{alpha}\PY{o}{=}\PY{l+m+mf}{0.2}\PY{p}{)}\PY{p}{(}\PY{n}{x}\PY{p}{)}
        \PY{c+c1}{\PYZsh{} x=BatchNormalization(momentum=0.8)(x)}
        \PY{n}{x}\PY{o}{=}\PY{n}{AveragePooling2D}\PY{p}{(}\PY{p}{(}\PY{l+m+mi}{2}\PY{p}{,} \PY{l+m+mi}{2}\PY{p}{)}\PY{p}{,} \PY{n}{padding}\PY{o}{=}\PY{l+s+s1}{\PYZsq{}}\PY{l+s+s1}{same}\PY{l+s+s1}{\PYZsq{}}\PY{p}{)}\PY{p}{(}\PY{n}{x}\PY{p}{)}
        \PY{n}{x}\PY{o}{=}\PY{n}{Conv2D}\PY{p}{(}\PY{n}{depth}\PY{o}{*}\PY{l+m+mi}{2}\PY{p}{,} \PY{p}{(}\PY{l+m+mi}{3}\PY{p}{,} \PY{l+m+mi}{3}\PY{p}{)}\PY{p}{,} \PY{n}{padding}\PY{o}{=}\PY{l+s+s1}{\PYZsq{}}\PY{l+s+s1}{same}\PY{l+s+s1}{\PYZsq{}}\PY{p}{)}\PY{p}{(}\PY{n}{x}\PY{p}{)}
        \PY{n}{x}\PY{o}{=}\PY{n}{LeakyReLU}\PY{p}{(}\PY{n}{alpha}\PY{o}{=}\PY{l+m+mf}{0.2}\PY{p}{)}\PY{p}{(}\PY{n}{x}\PY{p}{)}
        \PY{c+c1}{\PYZsh{} x=BatchNormalization(momentum=0.8)(x)}
        \PY{n}{x}\PY{o}{=}\PY{n}{Conv2D}\PY{p}{(}\PY{n}{depth}\PY{o}{*}\PY{l+m+mi}{2}\PY{p}{,} \PY{p}{(}\PY{l+m+mi}{3}\PY{p}{,} \PY{l+m+mi}{3}\PY{p}{)}\PY{p}{,} \PY{n}{padding}\PY{o}{=}\PY{l+s+s1}{\PYZsq{}}\PY{l+s+s1}{same}\PY{l+s+s1}{\PYZsq{}}\PY{p}{)}\PY{p}{(}\PY{n}{x}\PY{p}{)}
        \PY{n}{x}\PY{o}{=}\PY{n}{LeakyReLU}\PY{p}{(}\PY{n}{alpha}\PY{o}{=}\PY{l+m+mf}{0.2}\PY{p}{)}\PY{p}{(}\PY{n}{x}\PY{p}{)}
        \PY{c+c1}{\PYZsh{} x=BatchNormalization(momentum=0.8)(x)}
        \PY{n}{x}\PY{o}{=}\PY{n}{AveragePooling2D}\PY{p}{(}\PY{p}{(}\PY{l+m+mi}{2}\PY{p}{,} \PY{l+m+mi}{2}\PY{p}{)}\PY{p}{,} \PY{n}{padding}\PY{o}{=}\PY{l+s+s1}{\PYZsq{}}\PY{l+s+s1}{same}\PY{l+s+s1}{\PYZsq{}}\PY{p}{)}\PY{p}{(}\PY{n}{x}\PY{p}{)}
        \PY{n}{x}\PY{o}{=}\PY{n}{Conv2D}\PY{p}{(}\PY{n}{depth}\PY{o}{*}\PY{l+m+mi}{4}\PY{p}{,} \PY{p}{(}\PY{l+m+mi}{3}\PY{p}{,} \PY{l+m+mi}{3}\PY{p}{)}\PY{p}{,} \PY{n}{padding}\PY{o}{=}\PY{l+s+s1}{\PYZsq{}}\PY{l+s+s1}{same}\PY{l+s+s1}{\PYZsq{}}\PY{p}{)}\PY{p}{(}\PY{n}{x}\PY{p}{)}
        \PY{n}{x}\PY{o}{=}\PY{n}{LeakyReLU}\PY{p}{(}\PY{n}{alpha}\PY{o}{=}\PY{l+m+mf}{0.2}\PY{p}{)}\PY{p}{(}\PY{n}{x}\PY{p}{)}
        \PY{c+c1}{\PYZsh{} x=BatchNormalization(momentum=0.8)(x)}
        \PY{n}{x}\PY{o}{=}\PY{n}{Conv2D}\PY{p}{(}\PY{n}{depth}\PY{o}{*}\PY{l+m+mi}{4}\PY{p}{,} \PY{p}{(}\PY{l+m+mi}{3}\PY{p}{,} \PY{l+m+mi}{3}\PY{p}{)}\PY{p}{,} \PY{n}{padding}\PY{o}{=}\PY{l+s+s1}{\PYZsq{}}\PY{l+s+s1}{same}\PY{l+s+s1}{\PYZsq{}}\PY{p}{)}\PY{p}{(}\PY{n}{x}\PY{p}{)}
        \PY{n}{x}\PY{o}{=}\PY{n}{LeakyReLU}\PY{p}{(}\PY{n}{alpha}\PY{o}{=}\PY{l+m+mf}{0.2}\PY{p}{)}\PY{p}{(}\PY{n}{x}\PY{p}{)}
        \PY{c+c1}{\PYZsh{} x=BatchNormalization(momentum=0.8)(x)}
        \PY{n}{x}\PY{o}{=}\PY{n}{AveragePooling2D}\PY{p}{(}\PY{p}{(}\PY{l+m+mi}{2}\PY{p}{,} \PY{l+m+mi}{2}\PY{p}{)}\PY{p}{,} \PY{n}{padding}\PY{o}{=}\PY{l+s+s1}{\PYZsq{}}\PY{l+s+s1}{same}\PY{l+s+s1}{\PYZsq{}}\PY{p}{)}\PY{p}{(}\PY{n}{x}\PY{p}{)}
        \PY{n}{x}\PY{o}{=}\PY{n}{Flatten}\PY{p}{(}\PY{p}{)}\PY{p}{(}\PY{n}{x}\PY{p}{)}
        \PY{n}{x}\PY{o}{=}\PY{n}{Dense}\PY{p}{(}\PY{n}{interdim}\PY{p}{,}\PY{n}{activation}\PY{o}{=}\PY{l+s+s1}{\PYZsq{}}\PY{l+s+s1}{relu}\PY{l+s+s1}{\PYZsq{}}\PY{p}{)}\PY{p}{(}\PY{n}{x}\PY{p}{)}
        \PY{n}{encoded}\PY{o}{=}\PY{n}{Dense}\PY{p}{(}\PY{n}{endim}\PY{p}{)}\PY{p}{(}\PY{n}{x}\PY{p}{)}

        \PY{n}{encoder}\PY{o}{=}\PY{n}{Model}\PY{p}{(}\PY{n}{inputs}\PY{o}{=}\PY{p}{[}\PY{n}{img}\PY{p}{]}\PY{p}{,}\PY{n}{outputs}\PY{o}{=}\PY{p}{[}\PY{n}{encoded}\PY{p}{]}\PY{p}{)}
        \PY{n}{encoder}\PY{o}{.}\PY{n}{summary}\PY{p}{(}\PY{p}{)}
\end{Verbatim}

    \begin{Verbatim}[commandchars=\\\{\}]
\_\_\_\_\_\_\_\_\_\_\_\_\_\_\_\_\_\_\_\_\_\_\_\_\_\_\_\_\_\_\_\_\_\_\_\_\_\_\_\_\_\_\_\_\_\_\_\_\_\_\_\_\_\_\_\_\_\_\_\_\_\_\_\_\_
Layer (type)                 Output Shape              Param \#
=================================================================
input\_1 (InputLayer)         (None, 28, 28, 1)         0
\_\_\_\_\_\_\_\_\_\_\_\_\_\_\_\_\_\_\_\_\_\_\_\_\_\_\_\_\_\_\_\_\_\_\_\_\_\_\_\_\_\_\_\_\_\_\_\_\_\_\_\_\_\_\_\_\_\_\_\_\_\_\_\_\_
conv2d\_1 (Conv2D)            (None, 28, 28, 16)        160
\_\_\_\_\_\_\_\_\_\_\_\_\_\_\_\_\_\_\_\_\_\_\_\_\_\_\_\_\_\_\_\_\_\_\_\_\_\_\_\_\_\_\_\_\_\_\_\_\_\_\_\_\_\_\_\_\_\_\_\_\_\_\_\_\_
leaky\_re\_lu\_1 (LeakyReLU)    (None, 28, 28, 16)        0
\_\_\_\_\_\_\_\_\_\_\_\_\_\_\_\_\_\_\_\_\_\_\_\_\_\_\_\_\_\_\_\_\_\_\_\_\_\_\_\_\_\_\_\_\_\_\_\_\_\_\_\_\_\_\_\_\_\_\_\_\_\_\_\_\_
conv2d\_2 (Conv2D)            (None, 28, 28, 16)        2320
\_\_\_\_\_\_\_\_\_\_\_\_\_\_\_\_\_\_\_\_\_\_\_\_\_\_\_\_\_\_\_\_\_\_\_\_\_\_\_\_\_\_\_\_\_\_\_\_\_\_\_\_\_\_\_\_\_\_\_\_\_\_\_\_\_
leaky\_re\_lu\_2 (LeakyReLU)    (None, 28, 28, 16)        0
\_\_\_\_\_\_\_\_\_\_\_\_\_\_\_\_\_\_\_\_\_\_\_\_\_\_\_\_\_\_\_\_\_\_\_\_\_\_\_\_\_\_\_\_\_\_\_\_\_\_\_\_\_\_\_\_\_\_\_\_\_\_\_\_\_
average\_pooling2d\_1 (Average (None, 14, 14, 16)        0
\_\_\_\_\_\_\_\_\_\_\_\_\_\_\_\_\_\_\_\_\_\_\_\_\_\_\_\_\_\_\_\_\_\_\_\_\_\_\_\_\_\_\_\_\_\_\_\_\_\_\_\_\_\_\_\_\_\_\_\_\_\_\_\_\_
conv2d\_3 (Conv2D)            (None, 14, 14, 32)        4640
\_\_\_\_\_\_\_\_\_\_\_\_\_\_\_\_\_\_\_\_\_\_\_\_\_\_\_\_\_\_\_\_\_\_\_\_\_\_\_\_\_\_\_\_\_\_\_\_\_\_\_\_\_\_\_\_\_\_\_\_\_\_\_\_\_
leaky\_re\_lu\_3 (LeakyReLU)    (None, 14, 14, 32)        0
\_\_\_\_\_\_\_\_\_\_\_\_\_\_\_\_\_\_\_\_\_\_\_\_\_\_\_\_\_\_\_\_\_\_\_\_\_\_\_\_\_\_\_\_\_\_\_\_\_\_\_\_\_\_\_\_\_\_\_\_\_\_\_\_\_
conv2d\_4 (Conv2D)            (None, 14, 14, 32)        9248
\_\_\_\_\_\_\_\_\_\_\_\_\_\_\_\_\_\_\_\_\_\_\_\_\_\_\_\_\_\_\_\_\_\_\_\_\_\_\_\_\_\_\_\_\_\_\_\_\_\_\_\_\_\_\_\_\_\_\_\_\_\_\_\_\_
leaky\_re\_lu\_4 (LeakyReLU)    (None, 14, 14, 32)        0
\_\_\_\_\_\_\_\_\_\_\_\_\_\_\_\_\_\_\_\_\_\_\_\_\_\_\_\_\_\_\_\_\_\_\_\_\_\_\_\_\_\_\_\_\_\_\_\_\_\_\_\_\_\_\_\_\_\_\_\_\_\_\_\_\_
average\_pooling2d\_2 (Average (None, 7, 7, 32)          0
\_\_\_\_\_\_\_\_\_\_\_\_\_\_\_\_\_\_\_\_\_\_\_\_\_\_\_\_\_\_\_\_\_\_\_\_\_\_\_\_\_\_\_\_\_\_\_\_\_\_\_\_\_\_\_\_\_\_\_\_\_\_\_\_\_
conv2d\_5 (Conv2D)            (None, 7, 7, 64)          18496
\_\_\_\_\_\_\_\_\_\_\_\_\_\_\_\_\_\_\_\_\_\_\_\_\_\_\_\_\_\_\_\_\_\_\_\_\_\_\_\_\_\_\_\_\_\_\_\_\_\_\_\_\_\_\_\_\_\_\_\_\_\_\_\_\_
leaky\_re\_lu\_5 (LeakyReLU)    (None, 7, 7, 64)          0
\_\_\_\_\_\_\_\_\_\_\_\_\_\_\_\_\_\_\_\_\_\_\_\_\_\_\_\_\_\_\_\_\_\_\_\_\_\_\_\_\_\_\_\_\_\_\_\_\_\_\_\_\_\_\_\_\_\_\_\_\_\_\_\_\_
conv2d\_6 (Conv2D)            (None, 7, 7, 64)          36928
\_\_\_\_\_\_\_\_\_\_\_\_\_\_\_\_\_\_\_\_\_\_\_\_\_\_\_\_\_\_\_\_\_\_\_\_\_\_\_\_\_\_\_\_\_\_\_\_\_\_\_\_\_\_\_\_\_\_\_\_\_\_\_\_\_
leaky\_re\_lu\_6 (LeakyReLU)    (None, 7, 7, 64)          0
\_\_\_\_\_\_\_\_\_\_\_\_\_\_\_\_\_\_\_\_\_\_\_\_\_\_\_\_\_\_\_\_\_\_\_\_\_\_\_\_\_\_\_\_\_\_\_\_\_\_\_\_\_\_\_\_\_\_\_\_\_\_\_\_\_
average\_pooling2d\_3 (Average (None, 4, 4, 64)          0
\_\_\_\_\_\_\_\_\_\_\_\_\_\_\_\_\_\_\_\_\_\_\_\_\_\_\_\_\_\_\_\_\_\_\_\_\_\_\_\_\_\_\_\_\_\_\_\_\_\_\_\_\_\_\_\_\_\_\_\_\_\_\_\_\_
flatten\_1 (Flatten)          (None, 1024)              0
\_\_\_\_\_\_\_\_\_\_\_\_\_\_\_\_\_\_\_\_\_\_\_\_\_\_\_\_\_\_\_\_\_\_\_\_\_\_\_\_\_\_\_\_\_\_\_\_\_\_\_\_\_\_\_\_\_\_\_\_\_\_\_\_\_
dense\_1 (Dense)              (None, 128)               131200
\_\_\_\_\_\_\_\_\_\_\_\_\_\_\_\_\_\_\_\_\_\_\_\_\_\_\_\_\_\_\_\_\_\_\_\_\_\_\_\_\_\_\_\_\_\_\_\_\_\_\_\_\_\_\_\_\_\_\_\_\_\_\_\_\_
dense\_2 (Dense)              (None, 2)                 258
=================================================================
Total params: 203,250
Trainable params: 203,250
Non-trainable params: 0
\_\_\_\_\_\_\_\_\_\_\_\_\_\_\_\_\_\_\_\_\_\_\_\_\_\_\_\_\_\_\_\_\_\_\_\_\_\_\_\_\_\_\_\_\_\_\_\_\_\_\_\_\_\_\_\_\_\_\_\_\_\_\_\_\_

    \end{Verbatim}

    \hypertarget{define-decoder}{%
\subsubsection{Define Decoder}\label{define-decoder}}

    \begin{Verbatim}[commandchars=\\\{\}]
{\color{incolor}In [{\color{incolor}5}]:} \PY{n}{x}\PY{o}{=}\PY{n}{Dense}\PY{p}{(}\PY{n}{interdim}\PY{p}{)}\PY{p}{(}\PY{n}{embedd}\PY{p}{)}
        \PY{n}{x}\PY{o}{=}\PY{n}{Dense}\PY{p}{(}\PY{n}{depth}\PY{o}{*}\PY{l+m+mi}{64}\PY{p}{,}\PY{n}{activation}\PY{o}{=}\PY{l+s+s1}{\PYZsq{}}\PY{l+s+s1}{relu}\PY{l+s+s1}{\PYZsq{}}\PY{p}{)}\PY{p}{(}\PY{n}{x}\PY{p}{)}
        \PY{c+c1}{\PYZsh{} x=BatchNormalization(momentum=0.8)(x)}
        \PY{n}{x}\PY{o}{=}\PY{n}{Reshape}\PY{p}{(}\PY{p}{(}\PY{l+m+mi}{4}\PY{p}{,}\PY{l+m+mi}{4}\PY{p}{,}\PY{l+m+mi}{4}\PY{o}{*}\PY{n}{depth}\PY{p}{)}\PY{p}{)}\PY{p}{(}\PY{n}{x}\PY{p}{)}
        \PY{n}{x}\PY{o}{=}\PY{n}{UpSampling2D}\PY{p}{(}\PY{p}{(}\PY{l+m+mi}{2}\PY{p}{,} \PY{l+m+mi}{2}\PY{p}{)}\PY{p}{)}\PY{p}{(}\PY{n}{x}\PY{p}{)}
        \PY{n}{x}\PY{o}{=}\PY{n}{Conv2D}\PY{p}{(}\PY{n}{depth}\PY{o}{*}\PY{l+m+mi}{4}\PY{p}{,} \PY{p}{(}\PY{l+m+mi}{3}\PY{p}{,} \PY{l+m+mi}{3}\PY{p}{)}\PY{p}{,} \PY{n}{padding}\PY{o}{=}\PY{l+s+s1}{\PYZsq{}}\PY{l+s+s1}{same}\PY{l+s+s1}{\PYZsq{}}\PY{p}{)}\PY{p}{(}\PY{n}{x}\PY{p}{)}
        \PY{n}{x}\PY{o}{=}\PY{n}{LeakyReLU}\PY{p}{(}\PY{n}{alpha}\PY{o}{=}\PY{l+m+mf}{0.2}\PY{p}{)}\PY{p}{(}\PY{n}{x}\PY{p}{)}
        \PY{c+c1}{\PYZsh{} x=BatchNormalization(momentum=0.8)(x)}
        \PY{n}{x}\PY{o}{=}\PY{n}{Conv2D}\PY{p}{(}\PY{n}{depth}\PY{o}{*}\PY{l+m+mi}{4}\PY{p}{,} \PY{p}{(}\PY{l+m+mi}{3}\PY{p}{,} \PY{l+m+mi}{3}\PY{p}{)}\PY{p}{,} \PY{n}{padding}\PY{o}{=}\PY{l+s+s1}{\PYZsq{}}\PY{l+s+s1}{same}\PY{l+s+s1}{\PYZsq{}}\PY{p}{)}\PY{p}{(}\PY{n}{x}\PY{p}{)}
        \PY{n}{x}\PY{o}{=}\PY{n}{LeakyReLU}\PY{p}{(}\PY{n}{alpha}\PY{o}{=}\PY{l+m+mf}{0.2}\PY{p}{)}\PY{p}{(}\PY{n}{x}\PY{p}{)}
        \PY{n}{x}\PY{o}{=}\PY{n}{UpSampling2D}\PY{p}{(}\PY{p}{(}\PY{l+m+mi}{2}\PY{p}{,} \PY{l+m+mi}{2}\PY{p}{)}\PY{p}{)}\PY{p}{(}\PY{n}{x}\PY{p}{)}
        \PY{n}{x}\PY{o}{=}\PY{n}{Conv2D}\PY{p}{(}\PY{n}{depth}\PY{o}{*}\PY{l+m+mi}{4}\PY{p}{,} \PY{p}{(}\PY{l+m+mi}{3}\PY{p}{,} \PY{l+m+mi}{3}\PY{p}{)}\PY{p}{,} \PY{n}{padding}\PY{o}{=}\PY{l+s+s1}{\PYZsq{}}\PY{l+s+s1}{valid}\PY{l+s+s1}{\PYZsq{}}\PY{p}{)}\PY{p}{(}\PY{n}{x}\PY{p}{)}
        \PY{n}{x}\PY{o}{=}\PY{n}{LeakyReLU}\PY{p}{(}\PY{n}{alpha}\PY{o}{=}\PY{l+m+mf}{0.2}\PY{p}{)}\PY{p}{(}\PY{n}{x}\PY{p}{)}
        \PY{c+c1}{\PYZsh{} x=BatchNormalization(momentum=0.8)(x)}
        \PY{n}{x}\PY{o}{=}\PY{n}{Conv2D}\PY{p}{(}\PY{n}{depth}\PY{o}{*}\PY{l+m+mi}{4}\PY{p}{,} \PY{p}{(}\PY{l+m+mi}{3}\PY{p}{,} \PY{l+m+mi}{3}\PY{p}{)}\PY{p}{,} \PY{n}{padding}\PY{o}{=}\PY{l+s+s1}{\PYZsq{}}\PY{l+s+s1}{same}\PY{l+s+s1}{\PYZsq{}}\PY{p}{)}\PY{p}{(}\PY{n}{x}\PY{p}{)}
        \PY{n}{x}\PY{o}{=}\PY{n}{LeakyReLU}\PY{p}{(}\PY{n}{alpha}\PY{o}{=}\PY{l+m+mf}{0.2}\PY{p}{)}\PY{p}{(}\PY{n}{x}\PY{p}{)}
        \PY{n}{x}\PY{o}{=}\PY{n}{UpSampling2D}\PY{p}{(}\PY{p}{(}\PY{l+m+mi}{2}\PY{p}{,} \PY{l+m+mi}{2}\PY{p}{)}\PY{p}{)}\PY{p}{(}\PY{n}{x}\PY{p}{)}
        \PY{n}{x}\PY{o}{=}\PY{n}{Conv2D}\PY{p}{(}\PY{n}{depth}\PY{o}{*}\PY{l+m+mi}{2}\PY{p}{,} \PY{p}{(}\PY{l+m+mi}{3}\PY{p}{,} \PY{l+m+mi}{3}\PY{p}{)}\PY{p}{,} \PY{n}{padding}\PY{o}{=}\PY{l+s+s1}{\PYZsq{}}\PY{l+s+s1}{same}\PY{l+s+s1}{\PYZsq{}}\PY{p}{)}\PY{p}{(}\PY{n}{x}\PY{p}{)}
        \PY{n}{x}\PY{o}{=}\PY{n}{LeakyReLU}\PY{p}{(}\PY{n}{alpha}\PY{o}{=}\PY{l+m+mf}{0.2}\PY{p}{)}\PY{p}{(}\PY{n}{x}\PY{p}{)}
        \PY{c+c1}{\PYZsh{} x=BatchNormalization(momentum=0.8)(x)}
        \PY{n}{x}\PY{o}{=}\PY{n}{Conv2D}\PY{p}{(}\PY{n}{depth}\PY{o}{*}\PY{l+m+mi}{2}\PY{p}{,} \PY{p}{(}\PY{l+m+mi}{3}\PY{p}{,} \PY{l+m+mi}{3}\PY{p}{)}\PY{p}{,} \PY{n}{padding}\PY{o}{=}\PY{l+s+s1}{\PYZsq{}}\PY{l+s+s1}{same}\PY{l+s+s1}{\PYZsq{}}\PY{p}{)}\PY{p}{(}\PY{n}{x}\PY{p}{)}
        \PY{n}{x}\PY{o}{=}\PY{n}{LeakyReLU}\PY{p}{(}\PY{n}{alpha}\PY{o}{=}\PY{l+m+mf}{0.2}\PY{p}{)}\PY{p}{(}\PY{n}{x}\PY{p}{)}
        \PY{c+c1}{\PYZsh{} x=BatchNormalization(momentum=0.8)(x)}
        \PY{c+c1}{\PYZsh{} x=BatchNormalization(momentum=0.8)(x)}
        \PY{n}{decoded}\PY{o}{=}\PY{n}{Conv2D}\PY{p}{(}\PY{l+m+mi}{1}\PY{p}{,} \PY{p}{(}\PY{l+m+mi}{3}\PY{p}{,} \PY{l+m+mi}{3}\PY{p}{)}\PY{p}{,} \PY{n}{padding}\PY{o}{=}\PY{l+s+s1}{\PYZsq{}}\PY{l+s+s1}{same}\PY{l+s+s1}{\PYZsq{}}\PY{p}{,}\PY{n}{activation}\PY{o}{=}\PY{l+s+s1}{\PYZsq{}}\PY{l+s+s1}{sigmoid}\PY{l+s+s1}{\PYZsq{}}\PY{p}{)}\PY{p}{(}\PY{n}{x}\PY{p}{)}

        \PY{n}{decoder}\PY{o}{=}\PY{n}{Model}\PY{p}{(}\PY{n}{inputs}\PY{o}{=}\PY{p}{[}\PY{n}{embedd}\PY{p}{]}\PY{p}{,}\PY{n}{outputs}\PY{o}{=}\PY{p}{[}\PY{n}{decoded}\PY{p}{]}\PY{p}{)}
        \PY{n}{decoder}\PY{o}{.}\PY{n}{summary}\PY{p}{(}\PY{p}{)}
\end{Verbatim}

    \begin{Verbatim}[commandchars=\\\{\}]
\_\_\_\_\_\_\_\_\_\_\_\_\_\_\_\_\_\_\_\_\_\_\_\_\_\_\_\_\_\_\_\_\_\_\_\_\_\_\_\_\_\_\_\_\_\_\_\_\_\_\_\_\_\_\_\_\_\_\_\_\_\_\_\_\_
Layer (type)                 Output Shape              Param \#
=================================================================
input\_2 (InputLayer)         (None, 2)                 0
\_\_\_\_\_\_\_\_\_\_\_\_\_\_\_\_\_\_\_\_\_\_\_\_\_\_\_\_\_\_\_\_\_\_\_\_\_\_\_\_\_\_\_\_\_\_\_\_\_\_\_\_\_\_\_\_\_\_\_\_\_\_\_\_\_
dense\_3 (Dense)              (None, 128)               384
\_\_\_\_\_\_\_\_\_\_\_\_\_\_\_\_\_\_\_\_\_\_\_\_\_\_\_\_\_\_\_\_\_\_\_\_\_\_\_\_\_\_\_\_\_\_\_\_\_\_\_\_\_\_\_\_\_\_\_\_\_\_\_\_\_
dense\_4 (Dense)              (None, 1024)              132096
\_\_\_\_\_\_\_\_\_\_\_\_\_\_\_\_\_\_\_\_\_\_\_\_\_\_\_\_\_\_\_\_\_\_\_\_\_\_\_\_\_\_\_\_\_\_\_\_\_\_\_\_\_\_\_\_\_\_\_\_\_\_\_\_\_
reshape\_1 (Reshape)          (None, 4, 4, 64)          0
\_\_\_\_\_\_\_\_\_\_\_\_\_\_\_\_\_\_\_\_\_\_\_\_\_\_\_\_\_\_\_\_\_\_\_\_\_\_\_\_\_\_\_\_\_\_\_\_\_\_\_\_\_\_\_\_\_\_\_\_\_\_\_\_\_
up\_sampling2d\_1 (UpSampling2 (None, 8, 8, 64)          0
\_\_\_\_\_\_\_\_\_\_\_\_\_\_\_\_\_\_\_\_\_\_\_\_\_\_\_\_\_\_\_\_\_\_\_\_\_\_\_\_\_\_\_\_\_\_\_\_\_\_\_\_\_\_\_\_\_\_\_\_\_\_\_\_\_
conv2d\_7 (Conv2D)            (None, 8, 8, 64)          36928
\_\_\_\_\_\_\_\_\_\_\_\_\_\_\_\_\_\_\_\_\_\_\_\_\_\_\_\_\_\_\_\_\_\_\_\_\_\_\_\_\_\_\_\_\_\_\_\_\_\_\_\_\_\_\_\_\_\_\_\_\_\_\_\_\_
leaky\_re\_lu\_7 (LeakyReLU)    (None, 8, 8, 64)          0
\_\_\_\_\_\_\_\_\_\_\_\_\_\_\_\_\_\_\_\_\_\_\_\_\_\_\_\_\_\_\_\_\_\_\_\_\_\_\_\_\_\_\_\_\_\_\_\_\_\_\_\_\_\_\_\_\_\_\_\_\_\_\_\_\_
conv2d\_8 (Conv2D)            (None, 8, 8, 64)          36928
\_\_\_\_\_\_\_\_\_\_\_\_\_\_\_\_\_\_\_\_\_\_\_\_\_\_\_\_\_\_\_\_\_\_\_\_\_\_\_\_\_\_\_\_\_\_\_\_\_\_\_\_\_\_\_\_\_\_\_\_\_\_\_\_\_
leaky\_re\_lu\_8 (LeakyReLU)    (None, 8, 8, 64)          0
\_\_\_\_\_\_\_\_\_\_\_\_\_\_\_\_\_\_\_\_\_\_\_\_\_\_\_\_\_\_\_\_\_\_\_\_\_\_\_\_\_\_\_\_\_\_\_\_\_\_\_\_\_\_\_\_\_\_\_\_\_\_\_\_\_
up\_sampling2d\_2 (UpSampling2 (None, 16, 16, 64)        0
\_\_\_\_\_\_\_\_\_\_\_\_\_\_\_\_\_\_\_\_\_\_\_\_\_\_\_\_\_\_\_\_\_\_\_\_\_\_\_\_\_\_\_\_\_\_\_\_\_\_\_\_\_\_\_\_\_\_\_\_\_\_\_\_\_
conv2d\_9 (Conv2D)            (None, 14, 14, 64)        36928
\_\_\_\_\_\_\_\_\_\_\_\_\_\_\_\_\_\_\_\_\_\_\_\_\_\_\_\_\_\_\_\_\_\_\_\_\_\_\_\_\_\_\_\_\_\_\_\_\_\_\_\_\_\_\_\_\_\_\_\_\_\_\_\_\_
leaky\_re\_lu\_9 (LeakyReLU)    (None, 14, 14, 64)        0
\_\_\_\_\_\_\_\_\_\_\_\_\_\_\_\_\_\_\_\_\_\_\_\_\_\_\_\_\_\_\_\_\_\_\_\_\_\_\_\_\_\_\_\_\_\_\_\_\_\_\_\_\_\_\_\_\_\_\_\_\_\_\_\_\_
conv2d\_10 (Conv2D)           (None, 14, 14, 64)        36928
\_\_\_\_\_\_\_\_\_\_\_\_\_\_\_\_\_\_\_\_\_\_\_\_\_\_\_\_\_\_\_\_\_\_\_\_\_\_\_\_\_\_\_\_\_\_\_\_\_\_\_\_\_\_\_\_\_\_\_\_\_\_\_\_\_
leaky\_re\_lu\_10 (LeakyReLU)   (None, 14, 14, 64)        0
\_\_\_\_\_\_\_\_\_\_\_\_\_\_\_\_\_\_\_\_\_\_\_\_\_\_\_\_\_\_\_\_\_\_\_\_\_\_\_\_\_\_\_\_\_\_\_\_\_\_\_\_\_\_\_\_\_\_\_\_\_\_\_\_\_
up\_sampling2d\_3 (UpSampling2 (None, 28, 28, 64)        0
\_\_\_\_\_\_\_\_\_\_\_\_\_\_\_\_\_\_\_\_\_\_\_\_\_\_\_\_\_\_\_\_\_\_\_\_\_\_\_\_\_\_\_\_\_\_\_\_\_\_\_\_\_\_\_\_\_\_\_\_\_\_\_\_\_
conv2d\_11 (Conv2D)           (None, 28, 28, 32)        18464
\_\_\_\_\_\_\_\_\_\_\_\_\_\_\_\_\_\_\_\_\_\_\_\_\_\_\_\_\_\_\_\_\_\_\_\_\_\_\_\_\_\_\_\_\_\_\_\_\_\_\_\_\_\_\_\_\_\_\_\_\_\_\_\_\_
leaky\_re\_lu\_11 (LeakyReLU)   (None, 28, 28, 32)        0
\_\_\_\_\_\_\_\_\_\_\_\_\_\_\_\_\_\_\_\_\_\_\_\_\_\_\_\_\_\_\_\_\_\_\_\_\_\_\_\_\_\_\_\_\_\_\_\_\_\_\_\_\_\_\_\_\_\_\_\_\_\_\_\_\_
conv2d\_12 (Conv2D)           (None, 28, 28, 32)        9248
\_\_\_\_\_\_\_\_\_\_\_\_\_\_\_\_\_\_\_\_\_\_\_\_\_\_\_\_\_\_\_\_\_\_\_\_\_\_\_\_\_\_\_\_\_\_\_\_\_\_\_\_\_\_\_\_\_\_\_\_\_\_\_\_\_
leaky\_re\_lu\_12 (LeakyReLU)   (None, 28, 28, 32)        0
\_\_\_\_\_\_\_\_\_\_\_\_\_\_\_\_\_\_\_\_\_\_\_\_\_\_\_\_\_\_\_\_\_\_\_\_\_\_\_\_\_\_\_\_\_\_\_\_\_\_\_\_\_\_\_\_\_\_\_\_\_\_\_\_\_
conv2d\_13 (Conv2D)           (None, 28, 28, 1)         289
=================================================================
Total params: 308,193
Trainable params: 308,193
Non-trainable params: 0
\_\_\_\_\_\_\_\_\_\_\_\_\_\_\_\_\_\_\_\_\_\_\_\_\_\_\_\_\_\_\_\_\_\_\_\_\_\_\_\_\_\_\_\_\_\_\_\_\_\_\_\_\_\_\_\_\_\_\_\_\_\_\_\_\_

    \end{Verbatim}

    Here we define Keras variables for \(\theta\) and sample \(z\)s.

    \begin{Verbatim}[commandchars=\\\{\}]

{\color{incolor}In [{\color{incolor}6}]:}  \PY{c+c1}{\PYZsh{}Define a Keras Variable for \PYZbs{}theta\PYZus{}ls}
	\PY{n}{theta}\PY{o}{=}\PY{n}{K}\PY{o}{.}\PY{n}{variable}\PY{p}{(}\PY{n}{generateTheta}\PY{p}{(}\PY{n}{L}\PY{p}{,}\PY{n}{endim}\PY{p}{)}\PY{p}{)}
         \PY{c+c1}{\PYZsh{}Define a Keras Variable for samples of z}
        \PY{n}{z}\PY{o}{=}\PY{n}{K}\PY{o}{.}\PY{n}{variable}\PY{p}{(}\PY{n}{generateZ}\PY{p}{(}\PY{n}{batchsize}\PY{p}{,}\PY{n}{endim}\PY{p}{)}\PY{p}{)}

\end{Verbatim}

    Put encoder and decoder together to get the autoencoder

    \begin{Verbatim}[commandchars=\\\{\}]
{\color{incolor}In [{\color{incolor}7}]:} \PY{c+c1}{\PYZsh{} Generate the autoencoder by combining encoder and decoder}
        \PY{n}{aencoded}\PY{o}{=}\PY{n}{encoder}\PY{p}{(}\PY{n}{img}\PY{p}{)}
        \PY{n}{ae}\PY{o}{=}\PY{n}{decoder}\PY{p}{(}\PY{n}{aencoded}\PY{p}{)}
        \PY{n}{autoencoder}\PY{o}{=}\PY{n}{Model}\PY{p}{(}\PY{n}{inputs}\PY{o}{=}\PY{p}{[}\PY{n}{img}\PY{p}{]}\PY{p}{,}\PY{n}{outputs}\PY{o}{=}\PY{p}{[}\PY{n}{ae}\PY{p}{]}\PY{p}{)}
        \PY{n}{autoencoder}\PY{o}{.}\PY{n}{summary}\PY{p}{(}\PY{p}{)}
\end{Verbatim}

    \begin{Verbatim}[commandchars=\\\{\}]
\_\_\_\_\_\_\_\_\_\_\_\_\_\_\_\_\_\_\_\_\_\_\_\_\_\_\_\_\_\_\_\_\_\_\_\_\_\_\_\_\_\_\_\_\_\_\_\_\_\_\_\_\_\_\_\_\_\_\_\_\_\_\_\_\_
Layer (type)                 Output Shape              Param \#
=================================================================
input\_1 (InputLayer)         (None, 28, 28, 1)         0
\_\_\_\_\_\_\_\_\_\_\_\_\_\_\_\_\_\_\_\_\_\_\_\_\_\_\_\_\_\_\_\_\_\_\_\_\_\_\_\_\_\_\_\_\_\_\_\_\_\_\_\_\_\_\_\_\_\_\_\_\_\_\_\_\_
model\_1 (Model)              (None, 2)                 203250
\_\_\_\_\_\_\_\_\_\_\_\_\_\_\_\_\_\_\_\_\_\_\_\_\_\_\_\_\_\_\_\_\_\_\_\_\_\_\_\_\_\_\_\_\_\_\_\_\_\_\_\_\_\_\_\_\_\_\_\_\_\_\_\_\_
model\_2 (Model)              (None, 28, 28, 1)         308193
=================================================================
Total params: 511,443
Trainable params: 511,443
Non-trainable params: 0
\_\_\_\_\_\_\_\_\_\_\_\_\_\_\_\_\_\_\_\_\_\_\_\_\_\_\_\_\_\_\_\_\_\_\_\_\_\_\_\_\_\_\_\_\_\_\_\_\_\_\_\_\_\_\_\_\_\_\_\_\_\_\_\_\_

    \end{Verbatim}

    \begin{Verbatim}[commandchars=\\\{\}]
{\color{incolor}In [{\color{incolor}8}]:} \PY{c+c1}{\PYZsh{} Let projae be the projection of the encoded samples}
        \PY{n}{projae}\PY{o}{=}\PY{n}{K}\PY{o}{.}\PY{n}{dot}\PY{p}{(}\PY{n}{aencoded}\PY{p}{,}\PY{n}{K}\PY{o}{.}\PY{n}{transpose}\PY{p}{(}\PY{n}{theta}\PY{p}{)}\PY{p}{)}
        \PY{c+c1}{\PYZsh{} Let projz be the projection of the \PYZdl{}q\PYZus{}Z\PYZdl{} samples}
        \PY{n}{projz}\PY{o}{=}\PY{n}{K}\PY{o}{.}\PY{n}{dot}\PY{p}{(}\PY{n}{z}\PY{p}{,}\PY{n}{K}\PY{o}{.}\PY{n}{transpose}\PY{p}{(}\PY{n}{theta}\PY{p}{)}\PY{p}{)}
        \PY{c+c1}{\PYZsh{} Calculate the Sliced Wasserstein distance by sorting }
        \PY{c+c1}{\PYZsh{} the projections and calculating the L2 distance between}
        \PY{n}{W2}\PY{o}{=}\PY{p}{(}\PY{n}{tf}\PY{o}{.}\PY{n}{nn}\PY{o}{.}\PY{n}{top\PYZus{}k}\PY{p}{(}\PY{n}{tf}\PY{o}{.}\PY{n}{transpose}\PY{p}{(}\PY{n}{projae}\PY{p}{)}\PY{p}{,}\PY{n}{k}\PY{o}{=}\PY{n}{batchsize}\PY{p}{)}\PY{o}{.}\PY{n}{values}\PY{o}{\PYZhy{}}
            \PY{n}{tf}\PY{o}{.}\PY{n}{nn}\PY{o}{.}\PY{n}{top\PYZus{}k}\PY{p}{(}\PY{n}{tf}\PY{o}{.}\PY{n}{transpose}\PY{p}{(}\PY{n}{projz}\PY{p}{)}\PY{p}{,}\PY{n}{k}\PY{o}{=}\PY{n}{batchsize}\PY{p}{)}\PY{o}{.}\PY{n}{values}\PY{p}{)}\PY{o}{*}\PY{o}{*}\PY{l+m+mi}{2}
\end{Verbatim}

    \begin{Verbatim}[commandchars=\\\{\}]
{\color{incolor}In [{\color{incolor}9}]:} \PY{n}{crossEntropyLoss}\PY{o}{=} \PY{p}{(}\PY{l+m+mf}{1.0}\PY{p}{)}\PY{o}{*}\PY{n}{K}\PY{o}{.}\PY{n}{mean}\PY{p}{(}\PY{n}{K}\PY{o}{.}\PY{n}{binary\PYZus{}crossentropy}\PY{p}{(}\PY{n}{K}\PY{o}{.}\PY{n}{flatten}\PY{p}{(}\PY{n}{img}\PY{p}{)}\PY{p}{,}
								\PY{n}{K}\PY{o}{.}\PY{n}{flatten}\PY{p}{(}\PY{n}{ae}\PY{p}{)}\PY{p}{)}\PY{p}{)}
        \PY{n}{L1Loss}\PY{o}{=} \PY{p}{(}\PY{l+m+mf}{1.0}\PY{p}{)}\PY{o}{*}\PY{n}{K}\PY{o}{.}\PY{n}{mean}\PY{p}{(}\PY{n}{K}\PY{o}{.}\PY{n}{abs}\PY{p}{(}\PY{n}{K}\PY{o}{.}\PY{n}{flatten}\PY{p}{(}\PY{n}{img}\PY{p}{)}\PY{o}{\PYZhy{}}\PY{n}{K}\PY{o}{.}\PY{n}{flatten}\PY{p}{(}\PY{n}{ae}\PY{p}{)}\PY{p}{)}\PY{p}{)}
        \PY{n}{W2Loss}\PY{o}{=} \PY{p}{(}\PY{l+m+mf}{10.0}\PY{p}{)}\PY{o}{*}\PY{n}{K}\PY{o}{.}\PY{n}{mean}\PY{p}{(}\PY{n}{W2}\PY{p}{)}
        \PY{c+c1}{\PYZsh{} I have a combination of L1 and Cross\PYZhy{}Entropy loss }
        \PY{c+c1}{\PYZsh{} for the first term and then and W2 for the second term}
        \PY{n}{vae\PYZus{}Loss}\PY{o}{=}\PY{n}{crossEntropyLoss}\PY{o}{+}\PY{n}{L1Loss}\PY{o}{+}\PY{n}{W2Loss}
        \PY{n}{autoencoder}\PY{o}{.}\PY{n}{add\PYZus{}loss}\PY{p}{(}\PY{n}{vae\PYZus{}Loss}\PY{p}{)} \PY{c+c1}{\PYZsh{} Add the custom loss to the model}
\end{Verbatim}

    \begin{Verbatim}[commandchars=\\\{\}]
{\color{incolor}In [{\color{incolor}10}]:} \PY{c+c1}{\PYZsh{}Compile the model}
         \PY{n}{autoencoder}\PY{o}{.}\PY{n}{compile}\PY{p}{(}\PY{n}{optimizer}\PY{o}{=}\PY{l+s+s1}{\PYZsq{}}\PY{l+s+s1}{rmsprop}\PY{l+s+s1}{\PYZsq{}}\PY{p}{,}\PY{n}{loss}\PY{o}{=}\PY{l+s+s1}{\PYZsq{}}\PY{l+s+s1}{\PYZsq{}}\PY{p}{)}
\end{Verbatim}

    \hypertarget{load-the-mnist-dataset}{%
\subsubsection{Load the MNIST dataset}\label{load-the-mnist-dataset}}

    \begin{Verbatim}[commandchars=\\\{\}]
{\color{incolor}In [{\color{incolor}11}]:} \PY{p}{(}\PY{n}{x\PYZus{}train}\PY{p}{,}\PY{n}{y\PYZus{}train}\PY{p}{)}\PY{p}{,}\PY{p}{(}\PY{n}{x\PYZus{}test}\PY{p}{,}\PY{n}{\PYZus{}}\PY{p}{)}\PY{o}{=}\PY{n}{mnist}\PY{o}{.}\PY{n}{load\PYZus{}data}\PY{p}{(}\PY{p}{)}
         \PY{n}{x\PYZus{}train}\PY{o}{=}\PY{n}{np}\PY{o}{.}\PY{n}{expand\PYZus{}dims}\PY{p}{(}\PY{n}{x\PYZus{}train}\PY{o}{.}\PY{n}{astype}\PY{p}{(}\PY{l+s+s1}{\PYZsq{}}\PY{l+s+s1}{float32}\PY{l+s+s1}{\PYZsq{}}\PY{p}{)}\PY{o}{/}\PY{l+m+mf}{255.}\PY{p}{,}\PY{l+m+mi}{3}\PY{p}{)}
\end{Verbatim}

    \begin{Verbatim}[commandchars=\\\{\}]
{\color{incolor}In [{\color{incolor}12}]:} \PY{n}{plt}\PY{o}{.}\PY{n}{imshow}\PY{p}{(}\PY{n}{np}\PY{o}{.}\PY{n}{squeeze}\PY{p}{(}\PY{n}{x\PYZus{}train}\PY{p}{[}\PY{l+m+mi}{0}\PY{p}{,}\PY{o}{.}\PY{o}{.}\PY{o}{.}\PY{p}{]}\PY{p}{)}\PY{p}{)}
         \PY{n}{plt}\PY{o}{.}\PY{n}{show}\PY{p}{(}\PY{p}{)}
\end{Verbatim}

    \begin{center}
    \adjustimage{max size={0.9\linewidth}{0.9\paperheight}}{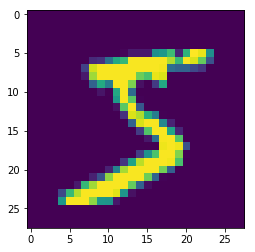}
    \end{center}
    { \hspace*{\fill} \\}

    \hypertarget{optimize-the-loss}{%
\subsection{Optimize the Loss}\label{optimize-the-loss}}

    \begin{Verbatim}[commandchars=\\\{\}]
{\color{incolor}In [{\color{incolor}13}]:} \PY{n}{loss}\PY{o}{=}\PY{p}{[}\PY{p}{]}
         \PY{k}{for} \PY{n}{epoch} \PY{o+ow}{in} \PY{n+nb}{range}\PY{p}{(}\PY{l+m+mi}{20}\PY{p}{)}\PY{p}{:}
             \PY{n}{ind}\PY{o}{=}\PY{n}{np}\PY{o}{.}\PY{n}{random}\PY{o}{.}\PY{n}{permutation}\PY{p}{(}\PY{n}{x\PYZus{}train}\PY{o}{.}\PY{n}{shape}\PY{p}{[}\PY{l+m+mi}{0}\PY{p}{]}\PY{p}{)}
             \PY{k}{for} \PY{n}{i} \PY{o+ow}{in} \PY{n+nb}{range}\PY{p}{(}\PY{n+nb}{int}\PY{p}{(}\PY{n}{x\PYZus{}train}\PY{o}{.}\PY{n}{shape}\PY{p}{[}\PY{l+m+mi}{0}\PY{p}{]}\PY{o}{/}\PY{n}{batchsize}\PY{p}{)}\PY{p}{)}\PY{p}{:}
                 \PY{n}{Xtr}\PY{o}{=}\PY{n}{x\PYZus{}train}\PY{p}{[}\PY{n}{ind}\PY{p}{[}\PY{n}{i}\PY{o}{*}\PY{n}{batchsize}\PY{p}{:}\PY{p}{(}\PY{n}{i}\PY{o}{+}\PY{l+m+mi}{1}\PY{p}{)}\PY{o}{*}\PY{n}{batchsize}\PY{p}{]}\PY{p}{,}\PY{o}{.}\PY{o}{.}\PY{o}{.}\PY{p}{]}
                 \PY{n}{theta\PYZus{}}\PY{o}{=}\PY{n}{generateTheta}\PY{p}{(}\PY{n}{L}\PY{p}{,}\PY{n}{endim}\PY{p}{)}
                 \PY{n}{z\PYZus{}}\PY{o}{=}\PY{n}{generateZ}\PY{p}{(}\PY{n}{batchsize}\PY{p}{,}\PY{n}{endim}\PY{p}{)}
                 \PY{n}{K}\PY{o}{.}\PY{n}{set\PYZus{}value}\PY{p}{(}\PY{n}{z}\PY{p}{,}\PY{n}{z\PYZus{}}\PY{p}{)}
                 \PY{n}{K}\PY{o}{.}\PY{n}{set\PYZus{}value}\PY{p}{(}\PY{n}{theta}\PY{p}{,}\PY{n}{theta\PYZus{}}\PY{p}{)}
                 \PY{n}{loss}\PY{o}{.}\PY{n}{append}\PY{p}{(}\PY{n}{autoencoder}\PY{o}{.}\PY{n}{train\PYZus{}on\PYZus{}batch}\PY{p}{(}\PY{n}{x}\PY{o}{=}\PY{n}{Xtr}\PY{p}{,}\PY{n}{y}\PY{o}{=}\PY{n+nb+bp}{None}\PY{p}{)}\PY{p}{)}
             \PY{n}{plt}\PY{o}{.}\PY{n}{plot}\PY{p}{(}\PY{n}{np}\PY{o}{.}\PY{n}{asarray}\PY{p}{(}\PY{n}{loss}\PY{p}{)}\PY{p}{)}
             \PY{n}{display}\PY{o}{.}\PY{n}{clear\PYZus{}output}\PY{p}{(}\PY{n}{wait}\PY{o}{=}\PY{n+nb+bp}{True}\PY{p}{)}
             \PY{n}{display}\PY{o}{.}\PY{n}{display}\PY{p}{(}\PY{n}{plt}\PY{o}{.}\PY{n}{gcf}\PY{p}{(}\PY{p}{)}\PY{p}{)}
             \PY{n}{time}\PY{o}{.}\PY{n}{sleep}\PY{p}{(}\PY{l+m+mf}{1e\PYZhy{}3}\PY{p}{)}

\end{Verbatim}

    \begin{center}
    \adjustimage{max size={0.9\linewidth}{0.9\paperheight}}{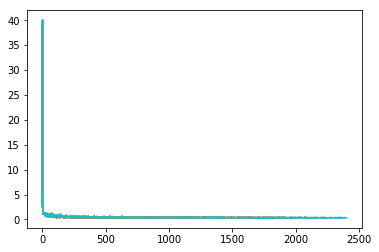}
    \end{center}
    { \hspace*{\fill} \\}

    \hypertarget{encode-and-decode-x_train}{%
\subsection{Encode and decode
x\_train}\label{encode-and-decode-x_train}}

    \begin{Verbatim}[commandchars=\\\{\}]
{\color{incolor}In [{\color{incolor}15}]:} \PY{c+c1}{\PYZsh{} Test autoencoder}
         \PY{n}{en}\PY{o}{=}\PY{n}{encoder}\PY{o}{.}\PY{n}{predict}\PY{p}{(}\PY{n}{x\PYZus{}train}\PY{p}{)}\PY{c+c1}{\PYZsh{} Encode the images}
         \PY{n}{dec}\PY{o}{=}\PY{n}{decoder}\PY{o}{.}\PY{n}{predict}\PY{p}{(}\PY{n}{en}\PY{p}{)} \PY{c+c1}{\PYZsh{} Decode the encodings}
\end{Verbatim}

    \begin{Verbatim}[commandchars=\\\{\}]
{\color{incolor}In [{\color{incolor}16}]:} \PY{c+c1}{\PYZsh{} Sanity check for the autoencoder}
         \PY{c+c1}{\PYZsh{} Note that we can use a more complex autoencoder that results }
         \PY{c+c1}{\PYZsh{} in better reconstructions. Also the autoencoders used in the  }
         \PY{c+c1}{\PYZsh{} literature often use a much larger latent space (we are using only 2!)}
         \PY{n}{fig}\PY{p}{,}\PY{p}{[}\PY{n}{ax1}\PY{p}{,}\PY{n}{ax2}\PY{p}{]}\PY{o}{=}\PY{n}{plt}\PY{o}{.}\PY{n}{subplots}\PY{p}{(}\PY{l+m+mi}{2}\PY{p}{,}\PY{l+m+mi}{1}\PY{p}{,}\PY{n}{figsize}\PY{o}{=}\PY{p}{(}\PY{l+m+mi}{100}\PY{p}{,}\PY{l+m+mi}{10}\PY{p}{)}\PY{p}{)}
         \PY{n}{I_temp}=\PY{p}{(}\PY{n}{stitchImages}\PY{p}{(}\PY{n}{x\PYZus{}train}\PY{p}{[}\PY{p}{:}\PY{l+m+mi}{10}\PY{p}{,}\PY{o}{.}\PY{o}{.}\PY{o}{.}\PY{p}{]}\PY{p}{,}\PY{l+m+mi}{1}\PY{p}{)}\PY{o}{*}\PY{l+m+mf}{255.0}\PY{p}{)}\PY{o}{.}\PY{n}{astype}\PY{p}{(}\PY{l+s+s1}{\PYZsq{}}\PY{l+s+s1}{uint8}\PY{l+s+s1}{\PYZsq{}}\PY{p}{)}
         \PY{n}{Idec_temp}=\PY{p}{(}\PY{n}{stitchImages}\PY{p}{(}\PY{n}{dec}\PY{p}{[}\PY{p}{:}\PY{l+m+mi}{10}\PY{p}{,}\PY{o}{.}\PY{o}{.}\PY{o}{.}\PY{p}{]}\PY{p}{,}\PY{l+m+mi}{1}\PY{p}{)}\PY{o}{*}\PY{l+m+mf}{255.0}\PY{p}{)}\PY{o}{.}\PY{n}{astype}\PY{p}{(}\PY{l+s+s1}{\PYZsq{}}\PY{l+s+s1}{uint8}\PY{l+s+s1}{\PYZsq{}}\PY{p}{)}
         \PY{n}{ax1}\PY{o}{.}\PY{n}{imshow}\PY{p}{(}\PY{n}{np}\PY{o}{.}\PY{n}{squeeze}\PY{p}{(}\PY{n}{I_temp}\PY{p}{)}\PY{p}{)}
         \PY{n}{ax1}\PY{o}{.}\PY{n}{set\PYZus{}xticks}\PY{p}{(}\PY{p}{[}\PY{p}{]}\PY{p}{)}
         \PY{n}{ax1}\PY{o}{.}\PY{n}{set\PYZus{}yticks}\PY{p}{(}\PY{p}{[}\PY{p}{]}\PY{p}{)}
         \PY{n}{ax2}\PY{o}{.}\PY{n}{imshow}\PY{p}{(}\PY{n}{np}\PY{o}{.}\PY{n}{squeeze}\PY{p}{(}\PY{n}{Idec_temp}\PY{p}{)}\PY{p}{)}
         \PY{n}{ax2}\PY{o}{.}\PY{n}{set\PYZus{}xticks}\PY{p}{(}\PY{p}{[}\PY{p}{]}\PY{p}{)}
         \PY{n}{ax2}\PY{o}{.}\PY{n}{set\PYZus{}yticks}\PY{p}{(}\PY{p}{[}\PY{p}{]}\PY{p}{)}
         \PY{n}{plt}\PY{o}{.}\PY{n}{show}\PY{p}{(}\PY{p}{)}
\end{Verbatim}

    \begin{center}
    \adjustimage{max size={0.9\linewidth}{0.9\paperheight}}{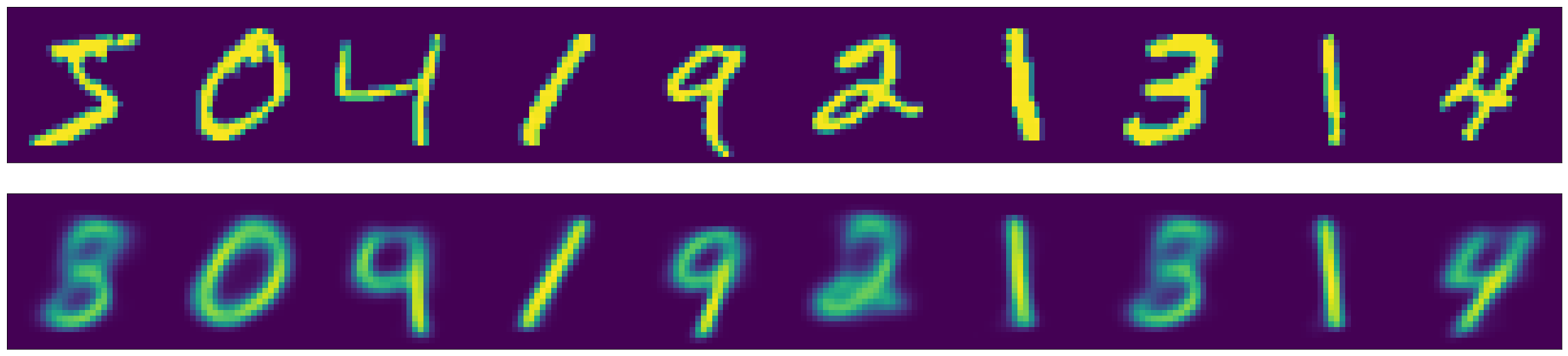}
    \end{center}
    { \hspace*{\fill} \\}

    \hypertarget{visualize-the-encoding-space}{%
\subsection{Visualize the encoding
space}\label{visualize-the-encoding-space}}

    \begin{Verbatim}[commandchars=\\\{\}]
{\color{incolor}In [{\color{incolor}17}]:} \PY{c+c1}{\PYZsh{} Distribution of the encoded samples}
         \PY{n}{plt}\PY{o}{.}\PY{n}{figure}\PY{p}{(}\PY{n}{figsize}\PY{o}{=}\PY{p}{(}\PY{l+m+mi}{10}\PY{p}{,}\PY{l+m+mi}{10}\PY{p}{)}\PY{p}{)}
         \PY{n}{plt}\PY{o}{.}\PY{n}{scatter}\PY{p}{(}\PY{n}{en}\PY{p}{[}\PY{p}{:}\PY{p}{,}\PY{l+m+mi}{0}\PY{p}{]}\PY{p}{,}\PY{o}{\PYZhy{}}\PY{n}{en}\PY{p}{[}\PY{p}{:}\PY{p}{,}\PY{l+m+mi}{1}\PY{p}{]}\PY{p}{,}\PY{n}{c}\PY{o}{=}\PY{l+m+mi}{10}\PY{o}{*}\PY{n}{y\PYZus{}train}\PY{p}{,} \PY{n}{cmap}\PY{o}{=}\PY{n}{plt}\PY{o}{.}\PY{n}{cm}\PY{o}{.}\PY{n}{Spectral}\PY{p}{)}
         \PY{n}{plt}\PY{o}{.}\PY{n}{xlim}\PY{p}{(}\PY{p}{[}\PY{o}{\PYZhy{}}\PY{l+m+mf}{1.5}\PY{p}{,}\PY{l+m+mf}{1.5}\PY{p}{]}\PY{p}{)}
         \PY{n}{plt}\PY{o}{.}\PY{n}{ylim}\PY{p}{(}\PY{p}{[}\PY{o}{\PYZhy{}}\PY{l+m+mf}{1.5}\PY{p}{,}\PY{l+m+mf}{1.5}\PY{p}{]}\PY{p}{)}
         \PY{n}{plt}\PY{o}{.}\PY{n}{show}\PY{p}{(}\PY{p}{)}
\end{Verbatim}

    \begin{center}
    \adjustimage{max size={0.9\linewidth}{0.9\paperheight}}{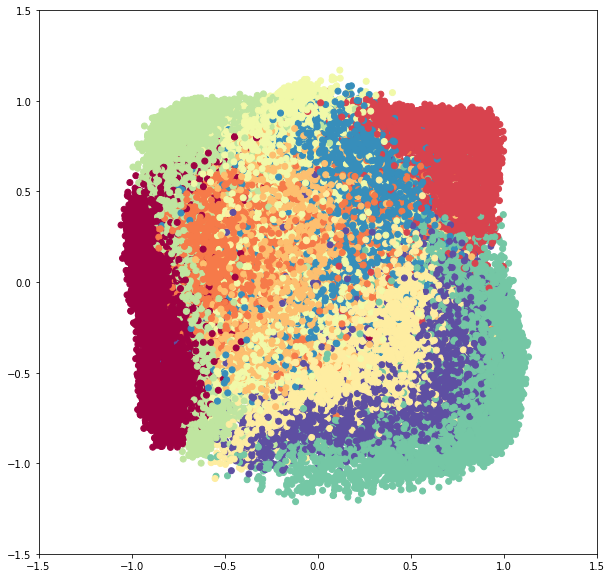}
    \end{center}
    { \hspace*{\fill} \\}

    \hypertarget{sample-a-grid-in-the-encoding-space-and-decode-it-to-visualize-this-space}{%
\subsubsection{Sample a grid in the encoding space and decode it to
visualize this
space}\label{sample-a-grid-in-the-encoding-space-and-decode-it-to-visualize-this-space}}

    \begin{Verbatim}[commandchars=\\\{\}]
{\color{incolor}In [{\color{incolor}18}]:} \PY{c+c1}{\PYZsh{}Sample the latent variable on a Nsample x Nsample grid}
         \PY{n}{Nsample}\PY{o}{=}\PY{l+m+mi}{25}
         \PY{n}{hiddenv}\PY{o}{=}\PY{n}{np}\PY{o}{.}\PY{n}{meshgrid}\PY{p}{(}\PY{n}{np}\PY{o}{.}\PY{n}{linspace}\PY{p}{(}\PY{o}{\PYZhy{}}\PY{l+m+mi}{1}\PY{p}{,}\PY{l+m+mi}{1}\PY{p}{,}\PY{n}{Nsample}\PY{p}{)}\PY{p}{,}\PY{n}{np}\PY{o}{.}\PY{n}{linspace}\PY{p}{(}\PY{o}{\PYZhy{}}\PY{l+m+mi}{1}\PY{p}{,}\PY{l+m+mi}{1}\PY{p}{,}\PY{n}{Nsample}\PY{p}{)}\PY{p}{)}
         \PY{n}{v}\PY{o}{=}\PY{n}{np}\PY{o}{.}\PY{n}{concatenate}\PY{p}{(}\PY{p}{(}\PY{n}{np}\PY{o}{.}\PY{n}{expand\PYZus{}dims}\PY{p}{(}\PY{n}{hiddenv}\PY{p}{[}\PY{l+m+mi}{0}\PY{p}{]}\PY{o}{.}\PY{n}{flatten}\PY{p}{(}\PY{p}{)}\PY{p}{,}\PY{l+m+mi}{1}\PY{p}{)}\PY{p}{,}
                           \PY{n}{np}\PY{o}{.}\PY{n}{expand\PYZus{}dims}\PY{p}{(}\PY{n}{hiddenv}\PY{p}{[}\PY{l+m+mi}{1}\PY{p}{]}\PY{o}{.}\PY{n}{flatten}\PY{p}{(}\PY{p}{)}\PY{p}{,}\PY{l+m+mi}{1}\PY{p}{)}\PY{p}{)}\PY{p}{,}\PY{l+m+mi}{1}\PY{p}{)}
         \PY{c+c1}{\PYZsh{} Decode the grid}
         \PY{n}{decodeimg}\PY{o}{=}\PY{n}{np}\PY{o}{.}\PY{n}{squeeze}\PY{p}{(}\PY{n}{decoder}\PY{o}{.}\PY{n}{predict}\PY{p}{(}\PY{n}{v}\PY{p}{)}\PY{p}{)}
\end{Verbatim}

    \begin{Verbatim}[commandchars=\\\{\}]
{\color{incolor}In [{\color{incolor}19}]:} \PY{c+c1}{\PYZsh{}Visualize the grid }
         \PY{n}{count}\PY{o}{=}\PY{l+m+mi}{0}
         \PY{n}{img}\PY{o}{=}\PY{n}{np}\PY{o}{.}\PY{n}{zeros}\PY{p}{(}\PY{p}{(}\PY{n}{Nsample}\PY{o}{*}\PY{l+m+mi}{28}\PY{p}{,}\PY{n}{Nsample}\PY{o}{*}\PY{l+m+mi}{28}\PY{p}{)}\PY{p}{)}
         \PY{k}{for} \PY{n}{i} \PY{o+ow}{in} \PY{n+nb}{range}\PY{p}{(}\PY{n}{Nsample}\PY{p}{)}\PY{p}{:}
             \PY{k}{for} \PY{n}{j} \PY{o+ow}{in} \PY{n+nb}{range}\PY{p}{(}\PY{n}{Nsample}\PY{p}{)}\PY{p}{:}
                 \PY{n}{img}\PY{p}{[}\PY{n}{i}\PY{o}{*}\PY{l+m+mi}{28}\PY{p}{:}\PY{p}{(}\PY{n}{i}\PY{o}{+}\PY{l+m+mi}{1}\PY{p}{)}\PY{o}{*}\PY{l+m+mi}{28}\PY{p}{,}\PY{n}{j}\PY{o}{*}\PY{l+m+mi}{28}\PY{p}{:}\PY{p}{(}\PY{n}{j}\PY{o}{+}\PY{l+m+mi}{1}\PY{p}{)}\PY{o}{*}\PY{l+m+mi}{28}\PY{p}{]}\PY{o}{=}\PY{n}{decodeimg}\PY{p}{[}\PY{n}{count}\PY{p}{,}\PY{o}{.}\PY{o}{.}\PY{o}{.}\PY{p}{]}
                 \PY{n}{count}\PY{o}{+}\PY{o}{=}\PY{l+m+mi}{1}
\end{Verbatim}

    \begin{Verbatim}[commandchars=\\\{\}]
{\color{incolor}In [{\color{incolor}20}]:} \PY{n}{fig}\PY{o}{=}\PY{n}{plt}\PY{o}{.}\PY{n}{figure}\PY{p}{(}\PY{n}{figsize}\PY{o}{=}\PY{p}{(}\PY{l+m+mi}{10}\PY{p}{,}\PY{l+m+mi}{10}\PY{p}{)}\PY{p}{)}
         \PY{n}{plt}\PY{o}{.}\PY{n}{imshow}\PY{p}{(}\PY{n}{img}\PY{p}{)}
         \PY{n}{plt}\PY{o}{.}\PY{n}{show}\PY{p}{(}\PY{p}{)}
\end{Verbatim}

    \begin{center}
    \adjustimage{max size={0.9\linewidth}{0.9\paperheight}}{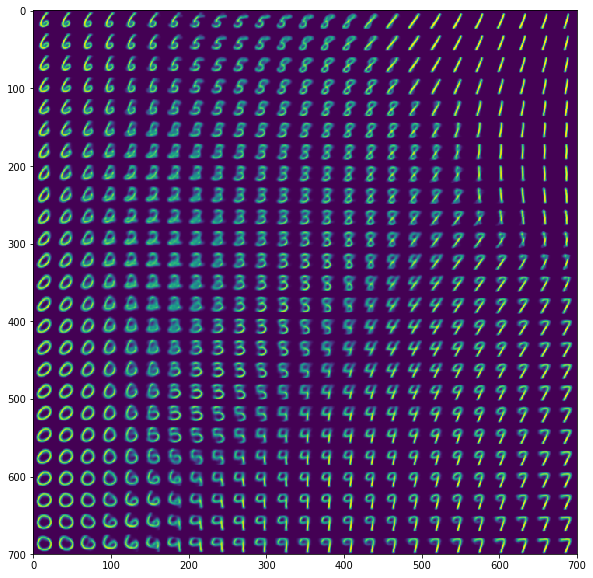}
    \end{center}
    { \hspace*{\fill} \\}

    \begin{Verbatim}[commandchars=\\\{\}]
{\color{incolor}In [{\color{incolor}21}]:} \PY{c+c1}{\PYZsh{}Visualize the z samples}
         \PY{n}{plt}\PY{o}{.}\PY{n}{figure}\PY{p}{(}\PY{n}{figsize}\PY{o}{=}\PY{p}{(}\PY{l+m+mi}{10}\PY{p}{,}\PY{l+m+mi}{10}\PY{p}{)}\PY{p}{)}
         \PY{n}{Z}\PY{o}{=}\PY{n}{generateZ}\PY{p}{(}\PY{l+m+mi}{10000}\PY{p}{,}\PY{l+m+mi}{2}\PY{p}{)}
         \PY{n}{plt}\PY{o}{.}\PY{n}{scatter}\PY{p}{(}\PY{n}{Z}\PY{p}{[}\PY{p}{:}\PY{p}{,}\PY{l+m+mi}{0}\PY{p}{]}\PY{p}{,}\PY{n}{Z}\PY{p}{[}\PY{p}{:}\PY{p}{,}\PY{l+m+mi}{1}\PY{p}{]}\PY{p}{)}
         \PY{n}{plt}\PY{o}{.}\PY{n}{xlim}\PY{p}{(}\PY{p}{[}\PY{o}{\PYZhy{}}\PY{l+m+mf}{1.5}\PY{p}{,}\PY{l+m+mf}{1.5}\PY{p}{]}\PY{p}{)}
         \PY{n}{plt}\PY{o}{.}\PY{n}{ylim}\PY{p}{(}\PY{p}{[}\PY{o}{\PYZhy{}}\PY{l+m+mf}{1.5}\PY{p}{,}\PY{l+m+mf}{1.5}\PY{p}{]}\PY{p}{)}
         \PY{n}{plt}\PY{o}{.}\PY{n}{show}\PY{p}{(}\PY{p}{)}
\end{Verbatim}

    \begin{center}
    \adjustimage{max size={0.9\linewidth}{0.9\paperheight}}{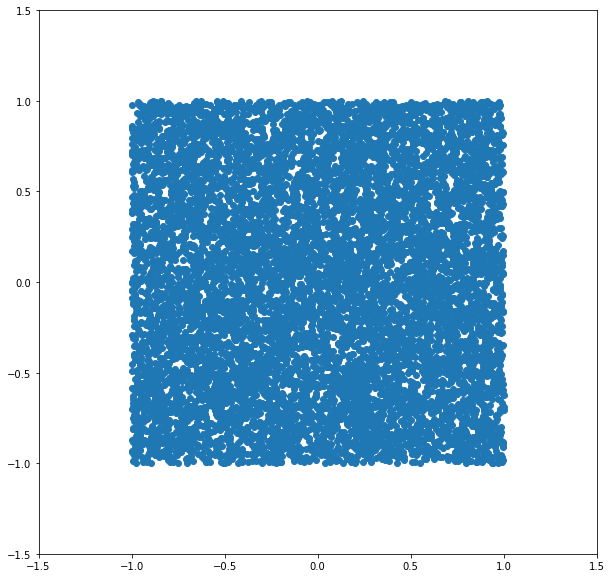}
    \end{center}
    { \hspace*{\fill} \\}

